\documentclass[11pt]{article}

\usepackage{graphicx}
\usepackage{latexsym}
\usepackage{amsmath}
\usepackage{amssymb}
\usepackage{natbib}
\usepackage{url}
\usepackage{amsthm}

\newtheorem{theorem}{Theorem}
\newtheorem{corollary}{Corollary}
\newtheorem{proposition}{Proposition}
\newtheorem{remark}{Remark}

\newtheorem{lemma}{Lemma}[section]

\newcommand{\bA}{\mbox{\boldmath {$A$}}}

\newcommand{\bB}{\mbox{\boldmath {$B$}}}

\newcommand{\bh}{\mbox{\boldmath {$h$}}}
\newcommand{\bH}{\mbox{\boldmath {$H$}}}

\newcommand{\bI}{\mbox{\boldmath {$I$}}}

\newcommand{\bM}{\mbox{\boldmath {$M$}}}

\newcommand{\bO}{\mbox{\boldmath {$O$}}}

\newcommand{\bP}{\mbox{\boldmath {$P$}}}

\newcommand{\bS}{\mbox{\boldmath {$S$}}}

\newcommand{\bu}{\mbox{\boldmath {$u$}}}

\newcommand{\bx}{\mbox{\boldmath {$x$}}}
\newcommand{\bX}{\mbox{\boldmath {$X$}}}
\newcommand{\by}{\mbox{\boldmath {$y$}}}

\newcommand{\bz}{\mbox{\boldmath {$z$}}}

\newcommand{\bze}{\mbox{\boldmath {$0$}}}
\newcommand{\bone}{\mbox{\boldmath {$1$}}}

\newcommand{\bmu}{\mbox{\boldmath $ \mu $}}
\newcommand{\bSig}{\mbox{\boldmath $ \Sigma $}}

\newcommand{\bSigma}{\mbox{\boldmath $ \Sigma $}}

\newcommand{\bLam}{\mbox{\boldmath $ \Lambda $}}

\newcommand{\bzeta}{\mbox{\boldmath $ \zeta $}}

\newcommand{\bOme}{\mbox{\boldmath $ \Omega $}}
\newcommand{\bGamma}{\mbox{\boldmath {$\Gamma$}}}

\newcommand{\tr}{\mbox{tr}}
\newcommand{\Var}{\mbox{Var}}

\usepackage{color}
\newcommand\CG[1]{\textcolor{black}{#1}}

\long\def\symbolfootnote[#1]#2{\begingroup%
\def\thefootnote{\fnsymbol{footnote}}\footnote[#1]{#2}\endgroup}

\begin{document}

\begin{center}
\Large
{\bf Distance-based classifier by data transformation for high-dimension, strongly spiked eigenvalue models}
\end{center}
\begin{center}
\vskip 0.5cm
\textbf{\large Makoto Aoshima and Kazuyoshi Yata} \\
Institute of Mathematics, University of Tsukuba, Ibaraki, Japan \\[-1cm]
\end{center}
\symbolfootnote[0]{\normalsize Address correspondence to Makoto Aoshima, 
Institute of Mathematics, University of Tsukuba, Ibaraki 305-8571, Japan; 
Fax: +81-298-53-6501; E-mail: aoshima@math.tsukuba.ac.jp}

\begin{abstract}
We consider classifiers for high-dimensional data under the strongly spiked eigenvalue (SSE) model. 
We first show that high-dimensional data often have the SSE model. 
We consider a distance-based classifier using eigenstructures for the SSE model.
We apply the noise reduction methodology to estimation of the eigenvalues and eigenvectors in the SSE model.
\CG{We create a new distance-based classifier by transforming data from the SSE model to the non-SSE model.} 
We give simulation studies and discuss the performance of the new classifier.
Finally, we demonstrate the new classifier by using microarray data sets.  
\end{abstract}

\section{Introduction}
\label{Sec1}
A common feature of high-dimensional data is that the data dimension is high, however, the sample size is relatively low. 
This is the so-called ``HDLSS" or ``large $p$, small $n$" data situation where $p/n\to\infty$; here $p$ is the data dimension and $n$ is the sample size. 
Suppose we have independent and $p$-variate two populations, $\pi_i,\ i=1,2$, having an unknown mean vector $\bmu_i$ and unknown covariance matrix $\bSigma_i$ 
for each $i$. 
We do not assume $\bSigma_1=\bSigma_2$.
The eigen-decomposition of $\bSigma_{i}$ is given by $\bSigma_{i}=\bH_{i}\bLam_{i}\bH_{i}^T$, where $\bLam_{i}=\mbox{diag}(\lambda_{i(1)},...,\lambda_{i(p)})$ is a diagonal matrix of eigenvalues, $\lambda_{i(1)}\ge \cdots \ge \lambda_{i(p)}\ge 0$, and $\bH_{i}=[\bh_{i(1)},...,\bh_{i(p)}]$ is an orthogonal matrix of the corresponding eigenvectors. 
We have independent and identically distributed (i.i.d.) observations, $\bx_{i1},...,\bx_{in_i}$, from each $\pi_i$.
We assume $n_i\ge 4,\ i=1,2$. 
We estimate $\bmu_i$ and $\bSigma_i$ by 
$\overline{\bx}_{i}=\sum_{j=1}^{n_i}{\bx_{ij}}/{n_i}$ 
and $\bS_{i}=\sum_{j=1}^{n_i}(\bx_{ij}-\overline{\bx}_{i})(\bx_{ij}-\overline{\bx}_{i})^T/(n_i-1)$.  
Let $\bx_0$ be an observation vector of an individual belonging to one of the two populations.
We assume $\bx_0$ and $\bx_{ij}$s are independent. 
\CG{When the $\pi_i$s are Gaussian}, 
a typical classification rule is that one classifies an individual into $\pi_1$ if
$$
(\bx_0-\overline{\bx}_{1})^T\bS_{1}^{-1}(\bx_0-\overline{\bx}_{1})-
\log\big\{\mbox{det}(\bS_{2}\bS_{1}^{-1} )\big\}
<
(\bx_0-\overline{\bx}_{2})^T\bS_{2}^{-1}(\bx_0-\overline{\bx}_{2}), 
$$
and into $\pi_2$ otherwise. 
However, the inverse matrix of $\bS_{i}$ does not exist in the HDLSS context ($p>n_i$). 
\CG{Also, we emphasize that the Gaussian assumption is strict in real high-dimensional data analyses.}  
\citet{Bickel:2004} \CG{considered a naive Bayes classifier for high-dimensional data}. 
\citet{Fan:2008} considered classification after feature selection. 
\citet{Cai:2011}, \citet{Shao:2011} and \citet{Li:2015} gave sparse linear or quadratic classification rules for high-dimensional data. 
The above references all assumed the following eigenvalues condition: 
There is a constant $c_{0}>0$ (not depending on $p$) such that 
\begin{equation}
c_{0}^{-1}< \lambda_{i(p)} \ \mbox{ and } \ \lambda_{i(1)}<c_{0} \ 
\mbox{ for $i=1,2$.}
\label{1.1}
\end{equation}
\citet{Dudoit:2002} considered using the inverse matrix defined by only diagonal elements of $\bS_{i}$.
\citet{Aoshima:2011,Aoshima:2015a} considered substituting $\{\tr(\bS_{i})/p\}\bI_p$ for $\bS_{i}$ by using the difference of a geometric representation of HDLSS data from each $\pi_i$. 
Here, $\bI_{p}$ denotes the identity matrix of dimension $p$. 
On the other hand, \citet{HMN:2005,Hall:2008} and \citet{Marron:2007} considered distance weighted classifiers. 
\citet{Ahn:2010} considered a HDLSS \CG{classifier based on the maximal} data piling.
\citet{HMN:2005}, \citet{Chan:2009}, \citet{Aoshima:2014} and \citet{Watanabe:2015} considered distance-based classifiers.
\citet{Aoshima:2014} gave the misclassification rate adjusted classifier for multiclass, high-dimensional data whose misclassification rates are no more than specified thresholds under the following eigenvalues condition: 
\begin{equation}
\frac{\lambda_{i(1)}^2}{ \tr(\bSigma_{i}^2)}\to 0 \ \mbox{ as $p\to \infty$ for $i=1,2$.} 
\label{NSSE}
\end{equation}
We emphasize that (\ref{NSSE}) is much milder than (\ref{1.1}) because (\ref{NSSE}) includes the case that $\lambda_{i(1)}\to \infty$ as $p\to \infty$. 
See Remark \ref{rem1} for the details.
\citet{Aoshima:2014} considered the distance-based classifier as follows: 
Let
\begin{align}
W(\bx_0)=\Big(\bx_0- \frac{  \overline{\bx}_{1}+ \overline{\bx}_{2}}{2} \Big)^T
(\overline{\bx}_{2}- \overline{\bx}_{1})-\frac{\tr(\bS_{1})}{2n_1}+\frac{\tr(\bS_{2})}{2n_2}.
\label{1.2}
\end{align}
Then, one classifies $\bx_0$ into $\pi_1$ if $W(\bx_0)<0$ and into $\pi_2$ otherwise. 
Here, $-\tr(\bS_{1})/(2n_1)+\tr(\bS_{2})/(2n_2)$ is a bias-correction term. 
Note that the classifier (\ref{1.2}) is equivalent to the scale adjusted distance-based classifier given by \citet{Chan:2009}. 
\CG{
\citet{Aoshima:2015b} called the classification rule (\ref{1.2}) the ``distance-based discriminant analysis (DBDA)". 
}

Recently, \citet{Aoshima:2016} considered the ``strongly spiked eigenvalue (SSE) model" as follows: 
\begin{equation}
\liminf_{p\to \infty}\Big\{ \frac{\lambda_{i(1)}^2}{\tr(\bSigma_{i}^2)}\Big\}>0\  
\label{SSE}
\mbox{ for $i=1$ or 2.}
\end{equation}
On the other hand, \citet{Aoshima:2016} called (\ref{NSSE}) the ``non-strongly spiked eigenvalue (NSSE) model". 
Note that 
(\ref{SSE}) holds under the condition:
\begin{equation}
\liminf_{p\to \infty}\Big\{ \frac{\lambda_{i(1)}}{\tr(\bSigma_{i})}\Big\}>0\ \mbox{ for $i=1$ or 2,}
\label{SSSE}
\end{equation}
from the fact that $\tr(\bSigma_{i}^2)\le \tr(\bSigma_{i})^2$. 
Here, $\lambda_{i(1)}/\tr(\bSigma_{i})$ is the first contribution ratio.
We call (\ref{SSSE}) the ``super strongly spiked eigenvalue (SSSE) model". 
\begin{remark}
\label{rem1}
Let us consider a spiked model such as 
\begin{equation}
\lambda_{i(r)}=a_{i(r)}p^{\alpha_{i(r)}}\ (r=1,...,t_i)\quad 
\mbox{and}\quad \lambda_{i(r)}=c_{i(r)}\ (r=t_i+1,...,p) \label{spiked}
\end{equation}
with positive and fixed constants, $a_{i(r)}$s, $c_{i(r)}$s and $\alpha_{i(r)}$s, and a positive and fixed integer $t_i$.
Note that \CG{the NSSE condition (\ref{NSSE})} holds when $\alpha_{i(1)}<1/2$ for $i=1,2$.
On the other hand, \CG{the SSE condition (\ref{SSE})} holds when $\alpha_{i(1)}\ge 1/2$, and further \CG{the SSSE condition (\ref{SSSE})} holds when $\alpha_{i(1)}\ge 1$.
See \citet{Yata:2012} for the details of the spiked model. 
\end{remark}

\begin{figure}
\includegraphics[scale=0.59]{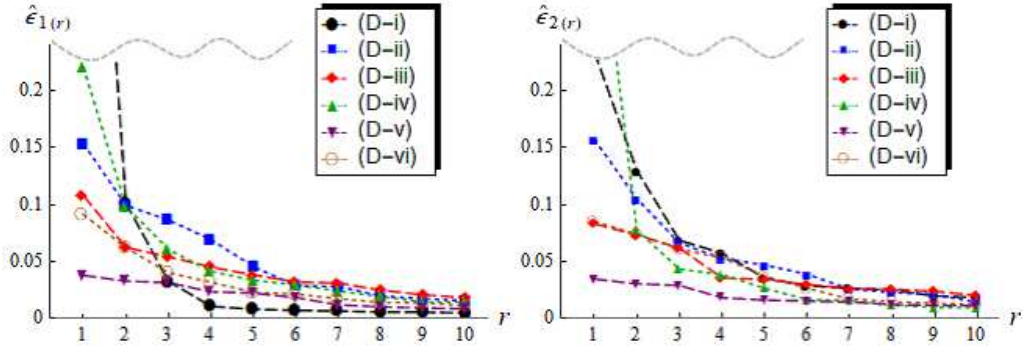}
\caption{
Estimates of the first ten contribution ratios  by $\hat{\varepsilon}_{i(r)}$s for the six well-known microarray data sets}
\label{fig:1} 
\end{figure}

\begin{figure}
\includegraphics[scale=0.59]{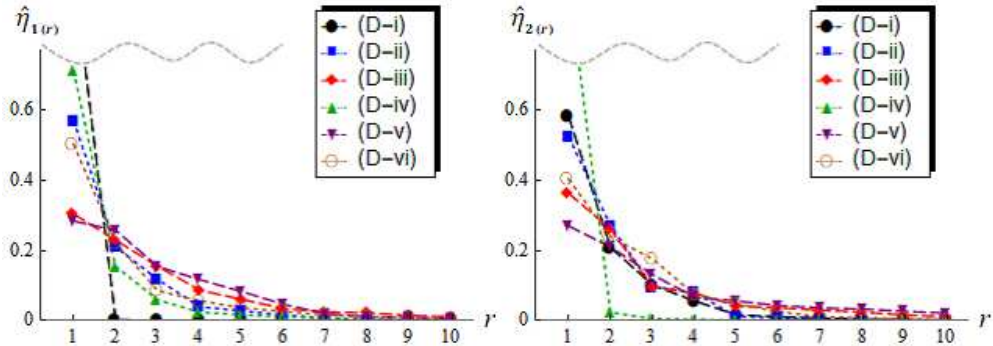}
\caption{
Estimates of the first ten quadratic contribution ratios by $\hat{\eta}_{i(r)}$s for the six well-known microarray data sets}
\label{fig:2} 
\end{figure}

We observed 
$$
\frac{\lambda_{i(r)}}{\tr(\bSigma_{i})}\ (=\varepsilon_{i(r)},\ \mbox{say}) \ \ 
\mbox{and} \ \ \frac{\lambda_{i(r)}^2}{\tr(\bSigma_{i}^2)}\ (=\eta_{i(r)},\ \mbox{say}), \ \ 
i=1,2;\ r=1,2,...,
$$
for six well-known microarray data sets by using the noise-reduction methodology and the cross-data-matrix methodology. 
For those methods, see \citet{Yata:2010,Yata:2012}. 
Note that $\varepsilon_{i(r)}$ is the contribution ratio and $\eta_{i(r)}$ is a quadratic contribution ratio of the $r$-th eigenvalue. 
We estimated $\varepsilon_{i(r)}$ by $\hat{\varepsilon}_{i(r)}= \tilde{\lambda}_{i(r)}/\tr(\bS_i)$ and 
$\eta_{i(r)}$ by $\hat{\eta}_{i(r)}=\acute{\lambda}_{i(r)}^2/\widehat{\Psi}_{i(1)}$, where 
$\tilde{\lambda}_{i(r)}$ is defined by (\ref{4.2}), and $\acute{\lambda}_{i(r)}$ and $\widehat{\Psi}_{i(1)}$ 
are defined in Section 4.3. 
We note that $\hat{\varepsilon}_{i(r)}$ and $\hat{\eta}_{i(r)}$ are consistent estimators of $\varepsilon_{i(r)}$ and $\eta_{i(r)}$ 
\CG{when $p\to \infty$.} 
See (\ref{nr}) and (\ref{cdm}) for the details. 
The six microarray data sets are as follows: 
\begin{description}
\item[(D-i)] \ 
\CG{
Non-pathologic tissues data with $1413$ genes, consisting of 
$\pi_1$: placenta or blood ($104$ samples) and $\pi_2:$ other solid tissue ($113$ samples) given by \citet{Christensen:2009};}
\item[(D-ii)] \ Colon cancer data with $2000$ genes, consisting of  
$\pi_1$: colon tumor ($40$ samples) and $\pi_2:$ normal colon ($22$ samples) given by \citet{Alon:1999};
\item[(D-iii)] \ 
\CG{
Breast cancer data with $2905$ genes, consisting of $\pi_1:$ good ($111$ samples) and $\pi_2:$ poor ($57$ samples) given by \citet{Gravier:2010};
}
\item[(D-iv)] \ Lymphoma data with $7129$ genes, consisting of $\pi_1:$ DLBCL (58 samples) and $\pi_2:$ follicular lymphoma (19 samples) given by \citet{Shipp:2002};
\item[(D-v)] \ Myeloma data with $12625$ genes, consisting of $\pi_1:$ 
patients without bone lesions (36 samples) and 
$\pi_2:$ patients with bone lesions (137 samples) given by \citet{Tian:2003}; 
\item[(D-vi)] \ Breast cancer data with $47293$ genes, consisting of $\pi_1:$ luminal group (84 samples) and $\pi_2:$ 
non-luminal group (44 samples) given by \citet{Naderi:2007}. 
\end{description}
The data sets (D-ii), (D-iv) and (D-v) are given in \citet{Jeffery:2006}, 
(D-i) and (D-iii) are given in \citet{Ramey:2016}, 
and (D-vi) is given in \citet{Glaab:2012}. 
We summarized the results for $\hat{\varepsilon}_{i(1)}$, $\hat{\eta}_{i(1)}$ \CG{and $\hat{k}_i$} in Table \ref{tab:1}, 
\CG{where $\hat{k}_i$ is an estimate of $k_i$, given in Section 4.3. 
We will discuss $k_i$ and $\hat{k}_i$ in Sections 3 and 4.3.}
We also visualized the first ten contribution ratios given by $\hat{\varepsilon}_{i(r)}\ (r=1,...,10;\ i=1,2)$ in Fig. \ref{fig:1} 
and the first ten quadratic contribution ratios given by $\hat{\eta}_{i(r)}\ (r=1,...,10;\ i=1,2)$ in Fig. \ref{fig:2}. See (\ref{nr}) and (\ref{cdm}) for the details.

\begin{table}
\caption{Estimates of ($\varepsilon_{i(1)}$,\ $\eta_{i(1)}$,\ $k_i$) by 
($\hat{\varepsilon}_{i(1)}$, $\hat{\eta}_{i(1)}$, $\hat{k}_i$) for the six well-known microarray data sets}
\label{tab:1}       
\begin{tabular}{ccccccc}
\hline
  & (D-i) & (D-ii) & (D-iii) & (D-iv) & (D-v) & (D-vi)  \\
\hline
$p$                  & 1413         & 2000 &2905    & 7129    &  12625   & 47293  \\
$(n_1,n_2)$         & (104,113)    &(40,22)&(111,57)& (58,19) & (36,137) & (84,44) \\
\hline
$\hat{\varepsilon}_{1(1)}$ &0.636  & 0.153 &0.108  & 0.22    & 0.038     &0.091   \\
$\hat{\varepsilon}_{2(1)}$ &0.233   & 0.157&0.083  & 0.386   & 0.035     &0.085  \\
\hline
$\hat{\eta}_{1(1)}$       &0.995   & 0.569 &0.304  & 0.71    &0.283      &0.502  \\
$\hat{\eta}_{2(1)}$       &0.582   & 0.523 &0.363  & 0.963   &0.269      &0.403  \\
\hline
\\[-3mm]
$\hat{k}_{1}$       &2   & 3 &2  & 2   &1      &2  \\
$\hat{k}_{2}$       &4   & 2 &2  & 2   &2      &3  \\
\hline
\end{tabular}
\end{table}

We observed from Fig. \ref{fig:1} that the first several eigenvalues are much larger than the rest for the microarray data sets (except (D-v)).
In particular, the first eigenvalues for (D-i) and (D-iv) are extremely large. 
\CG{These data appear to be consistent with the SSSE asymptotic domain given in (\ref{SSSE}).} 
On the other hand, the first several eigenvalues for (D-v) are relatively small. 
However, from Table \ref{tab:1} and Fig. \ref{fig:2}, $\eta_{i(1)}$s for (D-v) are not sufficiently small. 
\CG{
Also, \CG{$\eta_{i(1)}$s} 
for (D-ii), (D-iii) and (D-vi) are relatively large in Table \ref{tab:1} and Fig. \ref{fig:2}. 
Hence, the six microarray data appear to be consistent with the SSE asymptotic domain given in (\ref{SSE}).}  
See Section 4.3. 
In this paper, we consider classifiers under the SSE model. 
\CG{We do not assume the normality of the population distributions.} 
We propose an effective distance-based classifier for such high-dimensional data sets. 

The organization of this paper is as follows. 
In Section 2, we introduce asymptotic properties of the distance-based classifier for high-dimensional data.
We discuss the distance-based classifier in the SSE model. 
In Section 3, we consider a distance-based classifier using eigenstructures for the SSE model.
In Section 4, we discuss estimation of the eigenvalues and eigenvectors for the SSE model. 
We create a new distance-based classifier by estimating the eigenstructures. 
In Section 5, we give simulation studies and discuss the performance of the new classifier. 
Finally, we demonstrate the new classifier by using microarray data sets.  
\section{Distance-based classifier for high-dimensional data}
\label{Sec2}
In this section, we introduce asymptotic properties of the distance-based classifier for high-dimensional data. 
As for any positive-semidefinite matrix $\bM$, we write the square root of $\bM$ as $\bM^{1/2}$. 
Let 
$$
\bx_{ij}=\bH_i \bLam_i^{1/2}\bz_{ij}+\bmu_i, 
$$
where $\bz_{ij}=(z_{ij(1)},...,z_{ij(p)})^T$ is considered as a sphered data vector having the zero mean vector and identity covariance matrix. 
Similar to \citet{Bai:1996} and \citet{Chen:2010}, we assume the following assumption for $\pi_i$, $i=1,2$, 
as necessary: 
\begin{description}
\item[(A-i)] \ $\displaystyle \limsup_{p\to \infty} E(z_{ij(r)}^4)<\infty$ for all $r$, \ 
$E( z_{ij(r)}^2z_{ij(s)}^2)=E( z_{ij(r)}^2)E(z_{ij(s)}^2)=1$, \ $E( z_{ij(r)}z_{ij(s)}z_{ij(t)})=0$ \ and \
$E( z_{ij(r)} z_{ij(s)}z_{ij(t)}z_{ij(u)})=0$ for all $r \neq s,t,u$. 
\end{description} 
When the $\pi_i$s are Gaussian, (A-i) naturally holds. 
Let 
$$
\bmu=\bmu_1-\bmu_2,\ \ \Delta=\|\bmu\|^2, \ \ n_{\min}=\min\{n_1,n_2\} \ \mbox{ and } \ m=\min\{p,n_{\min} \},
$$
where $\|\cdot \|$ denotes the Euclidean norm. 
Note that $E\{W(\bx_0)\}=(-1)^i\Delta/2$ when $\bx_0\in\pi_i$ for $i=1,2$.
Also, note that the divergence condition ``$p\to \infty$, $n_1\to \infty$ and $n_2\to \infty$" is equivalent to ``$m\to \infty$".
Let 
$$
\delta_{oi}=
\Big\{ \frac{\tr(\bSigma_i^2)}{n_i}+
\frac{\tr(\bSigma_1\bSigma_2)}{n_{i'}}+
\sum_{l=1}^2\frac{\tr(\bSigma_{l}^2)}{2n_{l}(n_{l}-1)}
\Big\}^{1/2}
$$
and $\delta_i=\{\delta_{oi}^2+\bmu^T(\bSig_i+\bSig_{i'}/n_{i'})\bmu\}^{1/2}$ for $i=1,2;\ i'\neq i$.  
Note that 
$
\delta_i^2=\Var\{W(\bx_0)\}$ 
when $\bx_0\in\pi_i$ for $i=1,2$. 

\CG{Let $e(i)$ denote the error rate of misclassifying an individual from $\pi_i$ into the other class for $i=1,2$. 
Then, for the classification rule (\ref{1.2}) DBDA}, \citet{Aoshima:2014} gave the following result. 
\begin{theorem}[Aoshima and Yata, 2014]
\label{thm1}
Assume the following conditions: 
\begin{description}
\item[(AY-i)] \ $\displaystyle \frac{\bmu^T \bSigma_i\bmu}{ \Delta^2}\to 0$ \ as $p\to \infty$ for $i=1,2$;
\item[(AY-ii)] \ $\displaystyle \frac{\max_{i=1,2}\tr( \bSigma_i^2) }{n_{\min} \Delta^2}\to 0$ \ as $m\to \infty$. 
\end{description}
Then, for \CG{DBDA}, we have that as $m\to \infty$ 
\begin{align}
e(i)\to 0 \  \mbox{ for $i=1,2$}. \label{con}
\end{align}
\end{theorem}
\begin{remark}
\CG{
For DBDA, under (AY-i) and (AY-ii), 
one may write (\ref{con}) as }
$$
\CG{e(i)=O({\delta_i^2}/{\Delta^2} )\  \mbox{ for $i=1,2$}.}
$$
\end{remark}

Next, we consider the asymptotic normality of the classifier. 
\CG{Hereafter, for a function, $f(\cdot)$, ``$f(p) \in (0, \infty)$ as $p\to \infty$" implies $\liminf_{p\to \infty}f(p)>0$ and $\limsup_{p\to \infty}f(p)<\infty$.}
\CG{Let ``$\Rightarrow$" denote the convergence in distribution, $N(0,1)$ denote a random variable distributed as the standard normal distribution and $\Phi(\cdot )$ denote the cumulative distribution function of the standard normal distribution.}
\citet{Aoshima:2014} gave the following result. 
\begin{theorem}[Aoshima and Yata, 2014]
\label{thm2}
Assume the following conditions:
\begin{description}
\item[(AY-iii)] \ $\displaystyle \frac{\bmu^T \bSigma_i\bmu}{ \delta_{oi}^2 }\to 0$ \ as $m\to \infty$, \ \ 
$\displaystyle \liminf_{p\to \infty}\frac{ \tr(\bSigma_1\bSigma_2) }{ \tr(\bSigma_i^2)}>0$ for $i=1,2$, \ 
and \ $\displaystyle  \frac{ \tr(\bSigma_1^2) }{ \tr(\bSigma_2^2)}\in(0,\infty)$ as $p\to \infty$. 
\end{description}
Assume also \CG{the NSSE condition (\ref{NSSE})}.
Under a certain assumption milder than (A-i), it holds that as $m\to \infty$
$$
\frac{W(\bx_0)-(-1)^i\Delta/2}{\delta_{oi} }\Rightarrow N(0,1) \ \mbox{ when $\bx_0\in\pi_i$ for $i=1,2$}. 
$$
Furthermore, for \CG{DBDA}, it holds that as $m\to \infty$
\begin{align}
\label{asy}
e(i)- \Phi\Big( \frac{-\Delta}{2\delta_{oi}} \Big)=o(1) \ \mbox{ when $\bx_0\in\pi_i$ for $i=1,2$}.
\end{align}
\end{theorem}
\begin{remark}
\label{rem2}
\citet{Aoshima:2015b} gave a different asymptotic normality from Theorem \ref{thm2} under different conditions. 
From the facts that $\delta_{oi}/\delta_i\to 1$ as $m\to \infty$ under (AY-iii) and $\Var\{W(\bx_0)\}=\delta_i^2$ when $\bx_0\in\pi_i$,
one may write (\ref{asy}) as 
$$
e(i)- \Phi\{ -\Delta/(2\delta_{i}) \}=o(1) \ \mbox{ when $\bx_0\in\pi_i$ for $i=1,2$}.
$$
\end{remark}

By using the asymptotic normality, \citet{Aoshima:2014} proposed {\it the misclassification rate adjusted classifier (MRAC)} in high-dimensional settings.

In this paper, we consider the distance-based classifier from a different point of view.
We consider the classifier under the SSE model. 
We emphasize that high-dimensional data often have the SSE model. 
See Table \ref{tab:1}, Figs. \ref{fig:1} and \ref{fig:2}. 
If \CG{the SSE condition (\ref{SSE})} is met, one cannot claim the asymptotic normality in Theorem 2. 
In addition, if \CG{the SSE condition (\ref{SSE})} is met, (AY-ii) in Theorem 1 is equivalent to 
\begin{equation}
\lambda_{i(1)}^2/(n_{\min}\Delta^2)=o(1) \ \mbox{\ for $i=1,2.$}
\label{2.2}
\end{equation}
Thus (AY-ii) in the SSE model is stricter than that in the NSSE model, 
For example, for the NSSE model as 
the spiked model in (\ref{spiked}) with $\alpha_{i(1)}< 1/2$, $i=1,2$, 
(AY-ii) is equivalent to $p/(n_{\min}\Delta^2)=o(1)$.
On the other hand, for the SSE model as (\ref{spiked}) with $\alpha_{i(1)}>1/2$ (and $\alpha_{i(1)} \ge \alpha_{i'(1)}$ for $i'\neq i$), 
(AY-ii) is equivalent to $p^{2\alpha_{i(1)}}/(n_{\min}\Delta^2)=o(1)$. 
That means $n_{\min}$ or $\Delta$ should be quite large for the SSE model compared to the NSSE model. 
Thus if \CG{the SSE condition (\ref{SSE}) is met, DBDA 
has the classification consistency (\ref{con})} under strict conditions 
compared to \CG{the NSSE condition (\ref{NSSE})}.
In order to overcome the difficulties, 
we propose a new distance-based classifier by estimating eigenstructures for the SSE model. 
\section{Distance-based classifier using eigenstructures}
\label{Sec3}
Let 
$$
\Psi_{i(r)}=\tr(\bSig_i^2)-\sum_{s=1}^{r-1}\lambda_{i(s)}^2
= \sum_{s=r}^p\lambda_{i(s)}^2\quad \mbox{for $i=1,2$; $r=1,...,p$.}
$$
In this section, similar to \citet{Aoshima:2016}, 
we assume the following model for $i=1,2$: 
\begin{description}
\item[(M-i)] \ There exists a fixed integer $k_i\ (\ge 1)$ such that $\lambda_{i(1)},...,\lambda_{i(k_i)}$ are distinct in the sense that $\liminf_{p\to \infty}(\lambda_{i(r)}/\lambda_{i(s)}-1)>0$ when $1\le r<s\le k_i$, and $\lambda_{i(k_i)}$ and $\lambda_{i(k_i+1)}$ satisfy 
$$
\liminf_{p\to \infty}\frac{\lambda_{ i(k_i)}^2}{\Psi_{i(k_i)} }>0
\ \ \mbox{and} \ \ \frac{\lambda_{i(k_i+1)}^2}{ \Psi_{i(k_i+1)} }\to 0 \ \mbox{ as $p\to \infty$}.
$$ 
\end{description}
Note that (M-i) implies 
\CG{the SSE condition (\ref{SSE})}, that is (M-i) is one of the SSE models. 
For example, 
(M-i) holds in the spiked model in (\ref{spiked}) with 
$$
\alpha_{i(1)}\ge \cdots \ge  \alpha_{i(k_i)}\ge 1/2>\alpha_{i(k_i+1)}\ge \cdots \ge \alpha_{i(t_i)} \ 
\mbox{ and } \ 
a_{i(r)}\neq a_{i(s)}
$$
for $1\le r< s\le k_i;\ i=1,2$. 
\CG{We emphasize that (M-i) is a natural model under the SSE condition (\ref{SSE}). 
See Fig. \ref{fig:2}.
The six microarray data appear to be consistent with (M-i).
Similar to (\ref{2.2}), we note that the sufficient condition (AY-ii) in Theorem 1 is equivalent to
\begin{equation*}
\sum_{r=1}^{k_i} \lambda_{i(r)}^2/(n_{\min}\Delta^2)=o(1) \ \mbox{\ for $i=1,2$}
\label{ssecon}
\end{equation*}
under (M-i). 
According to the arguments in the last paragraph of Section 2, if (M-i) is met, DBDA has 
the classification consistency (\ref{con}) under strict conditions 
compared to the NSSE condition (\ref{NSSE}).
Also, one cannot claim the asymptotic normality in Theorem 2 under (M-i). 
In order to overcome the difficulties, 
similar to \citet{Aoshima:2016}, we consider a data transformation from the SSE model to the NSSE model.}
\subsection{Data transformation}
\CG{Recall that $\bh_{i(r)}$ is the $r$-th eigenvector of $\bSig_i$}. 
Let 
$$
\bA_{i}=\bI_p-\sum_{r=1}^{k_i}\bh_{i(r)}\bh_{i(r)}^T=\sum_{r=k_i+1}^{p}\bh_{i(r)}\bh_{i(r)}^T \quad
\mbox{and} \quad \bx_{ij,A}=\bA_{i}\bx_{ij}
$$ 
for $j=1,...,n_i;\ i=1,2$. 
Note that $\bA_{i}^{2}=\bA_{i}$ for $i=1,2$. 
Let us write that $\bmu_{i,A}=\bA_{i} \bmu_i$, $\bSigma_{i,A}=\bA_i \bSig_i\bA_i=\sum_{r=k_i+1}^{p}\lambda_{i(r)} \bh_{i(r)}\bh_{i(r)}^T$, $i=1,2$, 
$\bmu_{A}=\bmu_{1,A}-\bmu_{2,A}$ and $\Delta_{A}=\|\bmu_{A}\|^2$. 
Note that $E(\bx_{ij,A})=\bmu_{i,A}$ and $\Var(\bx_{ij,A})=\bSigma_{i,A}$ for all $i,j$. 
Thus the transformed data, $\bx_{ij,A}$, has the NSSE model in the sense that 
$$
{\{\lambda_{\max}(\bSigma_{i,A})\}^2}/{\tr(\bSigma_{i,A}^2)}=
{\lambda_{i(k_i+1)}^2}/{\Psi_{i(k_i+1)}}\to 0 \ \mbox{ as $p\to \infty$},
$$
where $\lambda_{\max}(\bM)$ denotes the largest eigenvalue of any positive-semidefinite matrix, $\bM$.
\CG{Hence, we can say that a classifier by using the transformed data has the classification consistency (\ref{con}) 
under mild conditions compared to DBDA when (M-i) is met. 
In addition, one can claim the asymptotic normality of the classifier even when the SSE condition (\ref{SSE}) is met.} 

\CG{Now, we propose the classifier by using the transformed data.} 
Let us write that
$\bA_{*}=(\bA_{1}+\bA_{2})/2$, 
$\bx_{0,A*}=\bA_{*}\bx_0$ and $\overline{\bx}_{i,A}=\sum_{j=1}^{n_i}\bx_{ij,A}/n_i=\bA_{i}\overline{\bx}_{i}$ for $i=1,2$. 
We consider the following classifier: 
\begin{align}
W_{A}(\bx_0) 
&=\Big(\bx_{0,A*}- \frac{ \overline{\bx}_{1,A}+\overline{\bx}_{2,A}}{2} \Big)^T
(\overline{\bx}_{2,A}- \overline{\bx}_{1,A})-\frac{\tr(\bA_{1} \bS_{1})}{2n_1}+\frac{\tr(\bA_{2}\bS_{2})}{2n_2} \notag \\
&=\bx_{0,A*}^T(\overline{\bx}_{2,A}- \overline{\bx}_{1,A})+\sum_{ j<j' }^{n_1} \frac{\bx_{1j,A}^T \bx_{1j',A}}{n_1(n_1-1)}-
\sum_{j<j'}^{n_2} \frac{\bx_{2j,A}^T \bx_{2j',A}}{n_2(n_2-1)}. \label{3.1}
\end{align}
Then, one classifies $\bx_0$ into $\pi_1$ if $W_{A}(\bx_0)<0$ and into $\pi_2$ otherwise. 
Let $\bA_{1,2}=\bA_1-\bA_2$. 
Here, let us write that 
$\bSigma_{i,A*}=\bA_* \bSig_i\bA_*$, 
\begin{align*}
\delta_{oi,A}=&
\Big\{ \frac{\tr(\bSigma_{i,A*}\bSigma_{i,A})}{n_i}+
\frac{\tr(\bSigma_{i,A*} \bSigma_{i',A})}{n_{i'}}+
\sum_{l=1}^2\frac{\tr(\bSigma_{l,A}^2)}{2n_{l}(n_{l}-1)}
\Big\}^{1/2};\\
\mbox{and} \ \ 
\delta_{i,A}=&\Big\{\delta_{oi,A}^2+
\bmu_{A}^T\bSig_{i,A*} \bmu_{A}
+\bmu_i^T\bA_{1,2} \bSig_{i,A}\bA_{1,2}\bmu_i/(4n_i) \\  
& \ \ +
(\bmu_A-\bA_{1,2}\bmu_i/2)^T\bSig_{i',A}(\bmu_A-\bA_{1,2}\bmu_i/2)/n_{i'}\Big\}^{1/2}
\end{align*}
for $i=1,2;\ i'\neq i$. 
Then, we claim that when $\bx_0\in\pi_i$ for $i=1,2$,
\begin{equation}
E\{W_{A}(\bx_0)\}=(-1)^i\frac{\Delta_A}{2}
-(-1)^{i}\frac{\bmu_{i}^{T}\bA_{1,2} \bmu_A}{2}
\ \ \mbox{and} \ \ \Var\{W_{A}(\bx_0)\}=\delta_{i,A}^2.
\label{3.2}
\end{equation} 
\begin{remark}
\label{rem3}
In general, $\bmu_{i}^{T}\bA_{1,2}\bmu_A$ in (\ref{3.2}) is not sufficiently large because of rank$(\bA_{1,2}) \le k_1+k_2\ (<\infty) $.
If $\bA_{1}=\bA_{2}$, it holds that 
$E\{W_{A}(\bx_0)\}=(-1)^i\Delta_A/2$ and 
\begin{align}
\Var\{W_{A}(\bx_0)\}=&
 \frac{\tr(\bSigma_{i,A}^2)}{n_i}+
\frac{\tr(\bSigma_{1,A}\bSigma_{2,A})}{n_{i'}}+
\sum_{l=1}^2\frac{\tr(\bSigma_{l,A}^2)}{2n_{l}(n_{l}-1)}
\notag \\
&+
\bmu_{A}^T(\bSig_{i,A}+\bSig_{i',A}/n_{i'}) \bmu_{A} \notag
\end{align}
when $\bx_0\in\pi_i$ for $i=1,2;\ i'\neq i$.
\end{remark}

\CG{
In Sections 3.2 and 3.3, 
we give consistency properties and an asymptotic normality of $W_A(\bx_0)$. 
We assume the following conditions as necessary: }
\begin{description}
  \item[(C-i)] \ $\displaystyle \frac{\bmu_A^T (\bSigma_{i,A*}+\bSigma_{i',A}/n_{i'}) \bmu_A}{ \Delta_A^2}\to 0$ \ as $p\to \infty$ \ 
  for $i=1,2;\ i'\neq i$;
\item[(C-ii)] \ $\displaystyle \frac{\tr( \bSigma_{i,A*}\bSigma_{l,A}) }{n_{l} \Delta_A^2}\to 0$ \ as $m\to \infty$ \ for $i,l=1,2$;
\item[(C-iii)] \ $\displaystyle \frac{ \bmu_{i}^{T}\bA_{1,2}\bmu_A}{ \Delta_A}\to 0$ \ as $p\to \infty$ \ and \ 
  $\displaystyle \limsup_{m\to \infty} \frac{\bmu_{i}^T\bA_{1,2}^2 \bmu_{i}}{n_{\min}^{1/2} \Delta_A}<\infty$ \ for $i=1,2$;
  \item[(C-iv)] $\displaystyle \frac{\bmu_A^T (\bSigma_{i,A*}+\bSigma_{i',A}/n_{i'}) \bmu_A}{ \delta_{oi,A}^2}\to 0$ \ as $m\to \infty$,
   \ $\displaystyle  \liminf_{p\to \infty}\frac{\tr(\bSigma_{1,A}\bSigma_{2,A}) }{\tr(\bSigma_{i,A}^2)}> 0$ \ for $i=1,2\ (i'\neq i)$, \ 
and \ $\displaystyle  \frac{ \tr(\bSigma_{1,A}^2) }{ \tr(\bSigma_{2,A}^2)}\in(0,\infty)$ \ as $p\to \infty$;
  \item[(C-v)] $\displaystyle \frac{ \bmu_{i}^{T}\bA_{1,2}\bmu_A}{ \delta_{oi,A} }\to 0$ \ as $m\to \infty$, \ \ 
  $\displaystyle \limsup_{m\to \infty} \frac{\bmu_{i}^T\bA_{1,2}^2 \bmu_{i}}{n_{\min}^{1/2} \delta_{oi,A}}<\infty$, \\ 
and \ $\displaystyle \frac{\lambda_{\max}(\bSigma_{i,A*}^{1/2}  \bSig_{l,A}\bSigma_{i,A*}^{1/2}) }
{\tr(\bSigma_{i,A*}\bSigma_{l,A})}\to 0$ \ as $p\to \infty$ \ for $i,l=1,2$.
\end{description} 
\subsection{Consistency of the classifier (\ref{3.1})}
\label{Sec3.1}
We consider consistency properties of $W_{A}(\bx_0) $. 
\CG{We note that $\delta_{i,A}^2/\Delta_A^2\to 0$ as $m\to \infty$ under (C-i) to (C-iii). See Section 6.1.} 
Then, we have the following results. 
\begin{theorem}
\label{thm3}
Assume (M-i).
Assume also (C-i) to (C-iii). 
Then, it holds that as $m\to \infty$ 
$$
\frac{W_A(\bx_0)}{\Delta_A}=  \frac{(-1)^{i}}{2}+o_P(1) \ \mbox{ when $\bx_0\in\pi_i$ for $i=1,2$.}
$$
For the classification rule (\ref{3.1}), we have \CG{the classification consistency (\ref{con})} as $m\to \infty$.
\end{theorem}
\begin{corollary}
\label{cor1}
If $\bA_{1}=\bA_{2}$, for the classification rule (\ref{3.1}), we have \CG{the classification consistency (\ref{con})} as $m\to \infty$ under (M-i) 
and the following conditions: 
$$
\frac{\bmu_A^T \bSigma_{i,A}\bmu_A}{ \Delta_A^2}\to0 \ \mbox{ as $p\to \infty$ \ and } \ 
\frac{\tr( \bSigma_{i,A}^2) }{n_{\min} \Delta_A^2}\to 0 \ \mbox{ as $m\to \infty$ \ for $i=1,2$.}
$$
\end{corollary}
\begin{remark}
\CG{
For the classification rule (\ref{3.1}), under (M-i) and (C-i) to (C-iii),  
one may write (\ref{con}) as }
$$
\CG{e(i)=O({\delta_{i,A}^2}/{\Delta_A^2} )\  \mbox{ for $i=1,2$}.}
$$
\end{remark}

\CG{Now, we consider the sufficient condition (C-ii) in Theorem \ref{thm3}.} 
When 
$\lambda_{i(1)}^2$
$/\tr( \bSigma_{i,A}^2)\to \infty$ as $p\to \infty$ for $i=1,2$, it holds that 
$$
\tr(\bSigma_{i,A*}\bSigma_{l,A})\le \{\tr(\bSigma_{i,A*}^2)\tr(\bSigma_{l,A}^2) \}^{1/2}=o[\{\tr(\bSigma_{i}^2)\tr(\bSigma_{l}^2)\}^{1/2}]
$$ 
for $i,l=1,2$, from the fact that $\tr(\bSigma_{i,A*}^2)\le \tr(\bSigma_{i}^2)$. 
Then, (C-ii) is milder than (AY-ii) if $\Delta$ and $\Delta_{A}$ are of the same order. 
\subsection{Asymptotic normality of the classifier (\ref{3.1})}
\label{Sec3.2}
We consider the asymptotic normality of $W_A(\bx_0)$. 
\CG{We have the following results.} 
\begin{theorem}
\label{thm4}
Assume (A-i) and (M-i). 
Assume also (C-iv) and (C-v). 
Then, it holds that as $m\to \infty$ 
\begin{align}
\frac{W_A(\bx_0)-(-1)^i\Delta_{A}/2}{\delta_{oi,A} }\Rightarrow N(0,1) \ \mbox{ when $\bx_0\in\pi_i$ for $i=1,2$} \notag
\end{align}
Furthermore, for the classification rule (\ref{3.1}), it holds that as $m\to \infty$
\begin{align}
\label{asy2}
e(i)- \Phi\Big( \frac{-\Delta_{A}}{2\delta_{oi,A}} \Big)=o(1) \ \mbox{ when $\bx_0\in\pi_i$ for $i=1,2$}.
\end{align}
\end{theorem}
\begin{corollary}
\label{cor1}
If $\bA_{1}=\bA_{2}$, for the classification rule (\ref{3.1}), (\ref{asy2}) holds as $m\to \infty$ under (A-i), (M-i) 
and the following conditions: 
\begin{align*}
&\frac{\bmu_A^T \bSigma_{i,A} \bmu_A}{ \delta_{oi,A}^2}\to 0 \ \mbox{ as $m\to \infty$, } \ 
 \liminf_{p\to \infty}\frac{\tr(\bSigma_{1,A}\bSigma_{2,A}) }{\tr(\bSigma_{i,A}^2)}> 0 \ \mbox{ for $i=1,2$;}\\
&\mbox{and } \ \frac{ \tr(\bSigma_{1,A}^2) }{ \tr(\bSigma_{2,A}^2)}\in(0,\infty) \ \mbox{ as $p\to \infty$.} 
\end{align*}
\end{corollary}
\begin{remark}
\label{rem4}
From (\ref{A.4}) in Section 6, 
we note that $\delta_{oi,A}/\delta_{i,A}\to 1$ as $m\to \infty$ under (C-iv) and (C-v).
Hence, one may write (\ref{asy2}) as 
$$
e(i)- \Phi\{ -\Delta_{A}/(2\delta_{i,A}) \}=o(1) \ \mbox{ when $\bx_0\in\pi_i$ for $i=1,2$}.
$$
\end{remark}

\begin{figure}
\includegraphics[scale=0.55]{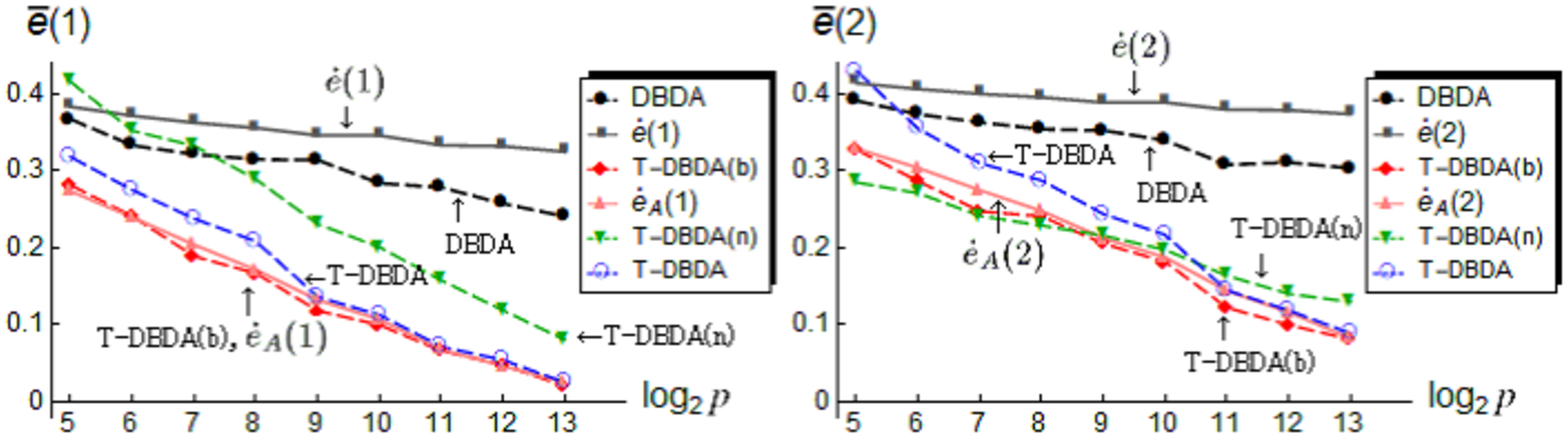}
\caption{
The left panel displays $\overline{e}(1)$ and the right panel displays $\overline{e}(2)$. 
The error rates (dashed lines) of DBDA (the classifier (\ref{1.2})), T-DBDA(b) (the classifier (\ref{3.1})), T-DBDA(n) (the classifier (\ref{3.5})) and T-DBDA (the classifier (\ref{new})). 
The asymptotic error rates (solid lines) by $\dot{e}(i)\ (=\Phi\{-\Delta/(2\delta_{i})\})$ and 
$\dot{e}_A(i)\ (=\Phi\{ -\Delta_{A}/(2\delta_{i,A}) \})$.}
\label{fig:3} 
\end{figure}

Now, let us show an easy example to check the performance of \CG{DBDA and the classifier (\ref{3.1})
for the SSE model.} 
We considered the following setting:
\begin{description}
\item[(S-i)] \ 
We set $p=2^{s},\ s=5,...,13$, and $n_1=\lceil p^{2/5} \rceil$ and $n_2=2n_1$, where $\lceil x \rceil$ denotes the smallest integer $\ge x$.   
Independent pseudo random observations were generated from $\pi_i: N_p(\bmu_i, \bSigma_i)$, $i=1,2$.
We set $\bmu_1=\bze$ and $\bmu_2=(0,...,0,1,...,1)^T$ whose last $\lceil p^{1/2} \rceil$ elements are $1$, 
$\bSig_1=\mbox{diag}(p^{2/3},p^{1/2},1,...,1)$ and $\bSig_2=2\bSig_1$. 
\end{description}
We note that (A-i), (M-i), (AY-i) to (AY-iii) and (C-i) to (C-v) are met for (S-i)
from the facts that $\Delta=\Delta_A=\lceil p^{1/2} \rceil$ and $\bA_{1}=\bA_{2}$ with $k_1=k_2=2$, 
so that Theorems \ref{thm1}, \ref{thm3} and \ref{thm4} hold.  
However, \CG{the NSSE condition (\ref{NSSE})} is not met, so that Theorem \ref{thm2} does not hold. 
In general, $\bA_{i}$s are unknown in (\ref{3.1}). 
Hence, we considered a naive estimator of $\bA_{i}$ as 
$
\widehat{\bA}_{i}=\bI_p-\sum_{r=1}^{k_i}\hat{\bh}_{i(r)}\hat{\bh}_{i(r)}^T
$
and checked the performance of the classifier given by 
\begin{align}
\widehat{W}_{A}(\bx_0) 
&=-\big\{\widehat{\bA}_{1}(\overline{\bx}_{1n_1}-\bx_0)+ \widehat{\bA}_{2}
(\overline{\bx}_{2n_2}-\bx_0)
 \big\}^T
\big(\widehat{\bA}_{2} \overline{\bx}_{2}
 - \widehat{\bA}_{1} \overline{\bx}_{1}\big)/2 \notag \\
& \quad -\tr(\widehat{\bA}_{1} \bS_{1})/(2n_1)+\tr(\widehat{\bA}_{2}\bS_{2})/(2n_2). \label{3.5}
\end{align}
Here, $\hat{\bh}_{i(r)}$ denotes the $r$-th (unit) eigenvector of $\bS_{i}$ for each $i,r$. 
Then, one classifies $\bx_0$ into $\pi_1$ if $\widehat{W}_{A}(\bx_0)<0$ and into $\pi_2$ otherwise. 
\CG{
On the other hand,  
by using a bias-corrected estimator of the eigenstructures, 
we create a new distance-based classifier given by (\ref{new}) in Section 4. 
We also checked the performance of the new classification rule (\ref{new}).  
We call the classification rule (\ref{new}) the ``transformed distance-based discriminant analysis (T-DBDA)". 
We also describe the classification rule (\ref{3.1}) as ``T-DBDA before estimation (T-DBDA(b))" and 
the classification rule (\ref{3.5}) as ``T-DBDA by the naive estimator (T-DBDA(n))".  
}
For $\bx_0\in\pi_i\ (i=1,2)$ we calculated each classifier 2000 times to confirm if each rule does (or does not) classify $\bx_0$ correctly and defined $P_{ir}=0\ (\mbox{or}\ 1)$ accordingly for each $\pi_i$.
We calculated the error rates, $\overline{e}(i)= \sum_{r=1}^{2000}P_{ir}/2000$, $i=1,2$. 
Their standard deviations are less than $0.011$. 
In Fig. \ref{fig:3}, we plotted $\overline{e}(1)$ and $\overline{e}(2)$ for \CG{DBDA, T-DBDA(n), T-DBDA(b) and T-DBDA}. 
From Theorems \ref{thm2} and \ref{thm4} in view of Remarks \ref{rem2} and \ref{rem4}, 
we also plotted the asymptotic error rates, 
\CG{
$\Phi\{-\Delta/(2\delta_{i})\}\ (=\dot{e}(i),\ \mbox{say})$ and $\Phi\{ -\Delta_{A}/(2\delta_{i,A}) \}\ (=\dot{e}_A(i),\ \mbox{say})$, in Fig. \ref{fig:3}.} 

We observed that $\overline{e}(i)$ by \CG{T-DBDA(b)} behaves very close to the asymptotic error rate, $\Phi\{ -\Delta_{A}/(2\delta_{i,A}) \}$, as expected theoretically. 
However, $\overline{e}(i)$ by \CG{DBDA} does not converge to $\Phi\{-\Delta/(2\delta_{i})\}$. 
This is because the classifier does not claim the asymptotic normality in Theorem \ref{thm2} for the SSE model. 
Both \CG{DBDA and T-DBDA(b) 
have the classification consistency (\ref{con})}. 
However, \CG{T-DBDA(b)} gave a much better performance compared to \CG{DBDA}. 
This is probably due to the convergence rates. 
For the sufficient conditions in Theorems \ref{thm1} and \ref{thm3}, we note that
\begin{align*}
&\max_{i=1,2} \tr( \bSigma_i^2)/(n_{\min} \Delta^2)=O(p^{1/3}/n_{\min})=O(p^{-1/15}) \ \mbox{ in (AY-ii);} \\
&\tr( \bSigma_{i,A*}\bSigma_{l,A})/(n_{l} \Delta_A^2)=O(n_l^{-1})=O(p^{-2/5}) \ \mbox{ for $i,l=1,2,$ \ in (C-ii)}.
\end{align*}
Hence, the error rates of \CG{T-DBDA(b)} were smaller than those of \CG{DBDA}. 
The \CG{T-DBDA(n)} gave a worse performance than \CG{T-DBDA(b)}.
This is probably because of the bias caused by the naive estimator, $\widehat{\bA}_{i}$. 
See Section 4.1 for the details. 
\CG{Hence, we will consider a bias-correction of the naive estimator in Section 4.  
On the other hand, the performances of \CG{T-DBDA and T-DBDA(b)} became similar to each other when $p$ is large. 
We will discuss T-DBDA in Section 4.2.} 

In Section 4, we discuss estimation of the unknown parameters in (\ref{3.1}). 
\CG{We create T-DBDA by the bias-corrected estimator of the parameters.}
\section{Distance-based classifier by estimating eigenstructures}
\label{Sec4}
In this section, we assume (A-i) and (M-i).
Let $x_{0,i(r)}=\bx_{0}^T{\bh}_{i(r)}$ and 
$$
x_{ij(r)}=\bx_{ij}^T{\bh}_{i(r)}=\lambda_{i(r)}^{1/2}z_{ij(r)}+\mu_{i(r)}\ \mbox{ for all $i,j,r$, where $\mu_{i(r)}=\bmu_i^T\bh_{i(r)}$}.
$$
Let us write that $\bar{x}_{i(r)}=\sum_{j=1}^{n_i} x_{ij(r)}/n_i$ for all $i,r$. 
Then, one can write (\ref{3.1}) as follows: 
\begin{align}
W_{A}(\bx_0)=&
W(\bx_0)+\sum_{r=1}^{k_1}x_{0,1(r)}\Big\{  \bar{x}_{1(r)}-\frac{1}{2}\bh_{1(r)}^T\Big( \overline{\bx}_{2}- \sum_{s=1}^{k_2}\bar{x}_{2(s)}  \bh_{2(s)}\Big) \Big\}
\notag \\
&-\sum_{r=1}^{k_2}x_{0,2(r)}\Big\{ \bar{x}_{2(r)}-\frac{1}{2} \bh_{2(r)}^T\Big( \overline{\bx}_{1}- \sum_{s=1}^{k_1}\bar{x}_{1(s)}  \bh_{1(s)}\Big)\Big\} \notag \\
&
-\sum_{r=1}^{k_1}\frac{\sum_{j<j'}^{n_1}x_{1j(r)}x_{1j'(r)}}{n_1(n_1-1)}+
 \sum_{r=1}^{k_2}\frac{\sum_{j<j'}^{n_2}x_{2j(r)}x_{2j'(r)} }{n_2(n_2-1)}.
 \label{4.1}
\end{align}
In order to use $W_{A}(\bx_0)$, it is necessary to estimate $\bh_{i(r)}$s, $x_{0,i(r)}$s, ${x}_{ij(r)}$s and $k_i$s. 

Let $\delta_{o\min ,A}=\min\{\delta_{o1,A},\delta_{o2,A} \}$. 
In this section, 
we assume the following conditions as necessary: 
\begin{description}
  \item[(C-vi)] \ 
$\displaystyle \limsup_{p\to \infty}\Big( \sum_{r=1}^{k_{i}}\frac{\bh_{i(r)}^T\bSig_{i'}\bh_{i(r)}}{ \lambda_{i(r)}}\Big)<\infty$ \ for $i=1,2\ (i'\neq i)$;
  \item[(C-vii)] \ 
  $\displaystyle \limsup_{m\to \infty}\Big(\sum_{r=1}^{k_i} \frac{n_{i}\{\mu_{i(r)}^2+(\bmu_{i'}^T\bh_{i(r)})^2\}}{\lambda_{i(r)}}\Big)<\infty$, \ \ $\displaystyle \limsup_{m\to \infty} \frac{\lambda_{l(1)} }{n_i \lambda_{i(k_i)}}<\infty$, \\
and \ $\displaystyle \limsup_{m\to \infty}\Big(\frac{\bmu_{l,A}^T\bSig_{i,A}\bmu_{l,A}}{\lambda_{i(k_i)}^2}\Big)<\infty$ 
 \ for $i,l=1,2\ (i'\neq i)$;
  \item[(C-viii)] \ 
  $\displaystyle  \frac{\lambda_{i(1)}}{n_{\min}\Delta_A }\to 0$ \ and \ 
  $\displaystyle  \frac{\bmu_{i,A}^T(\bSig_{i,A}/n_i+\bSig_{i',A}/n_{i'})\bmu_{i,A} }{\Delta_A^2}\to 0$ \ as $m\to \infty$ \ for $i=1,2\ (i'\neq i)$;
  \item[(C-ix)] \ 
    $\displaystyle  \frac{\lambda_{i(1)}}{n_{\min}\delta_{o\min,A} }\to 0$ \ and \ 
  $\displaystyle  \frac{\bmu_{i,A}^T(\bSig_{i,A}/n_i+\bSig_{i',A}/n_{i'})\bmu_{i,A} }{\delta_{o\min,A}^2}\to 0$ \ as $m\to \infty$ \ for $i=1,2\ (i'\neq i)$.
\end{description}
\subsection{Estimation of $\bh_{i(r)}$s, $x_{0,i(r)}$s and ${x}_{ij(r)}$s}
\label{Sec4.1}
Let $\bX_i=[\bx_{i1},...,\bx_{in}]$, $\overline{\bX}_i=[\overline{\bx}_{i},...,\overline{\bx}_{i}]$ 
and $\bP_{n_i}=\bI_{n_i}-\bone_{n_i}\bone_{n_i}^T/n_i$ for $i=1,2$, where $\bone_{n_i}=(1,...,1)^T$. 
Note that $\bS_{i}=\bX_i\bP_{n_i}\bX_i^T/(n_i-1)=(\bX_i-\overline{\bX}_i)(\bX_i-\overline{\bX}_i)^T/(n_i-1)$. 
We define the $n_i \times n_i$ dual sample covariance matrix by 
$$
\bS_{iD}=\bP_{n_i}\bX_i^T\bX_i\bP_{n_i}/(n_i-1)=(\bX_i-\overline{\bX}_i)^T(\bX_i-\overline{\bX}_i)/(n_i-1) \ \mbox{ for $i=1,2$}.
$$
Note that $\bS_i$ and $\bS_{iD}$ share non-zero eigenvalues. 
Let us write the eigen-decomposition of $\bS_{i}$ and $\bS_{iD}$ as
$$
\bS_{i}=\sum_{r=1}^{p}\hat{\lambda}_{i(r)}\hat{\bh}_{i(r)}\hat{\bh}_{i(r)}^T \ \mbox{ and }
 \ 
\bS_{iD}=\sum_{r=1}^{n_i-1}\hat{\lambda}_{i(r)}\hat{\bu}_{i(r)}\hat{\bu}_{i(r)}^T \ \mbox{ for $i=1,2$},
$$
where $\hat{\bh}_{i(r)}$ and $\hat{\bu}_{i(r)}$ denote unit eigenvectors corresponding to $\hat{\lambda}_{i(r)}$. 
We assume $\bh_{i(r)}^T\hat{\bh}_{i(r)} \ge 0$ w.p.1 for all $i,r$ without loss of generality. 
Note that $\hat{\bh}_{i(r)}$ can be calculated by 
$\hat{\bh}_{i(r)}=\{(n_i-1)\hat{\lambda}_{i(r)}\}^{-1/2}(\bX_i-\overline{\bX}_i) \hat{\bu}_{i(r)}$. 
However, as observed in Section 3.2, 
the classifier by $\hat{\bh}_{i(r)}$s gave an inadequate performance.

\citet{Yata:2012} proposed a bias-corrected eigenvalue estimation called the noise-reduction (NR) methodology, which was brought about by a geometric representation of $\bS_{iD}$.
If one applies the NR methodology, the $\lambda_{i(r)}$s are estimated by
\begin{equation}
\tilde{\lambda}_{i(r)}=\hat{\lambda}_{i(r)}-\frac{\tr(\bS_{iD})-\sum_{s=1}^r \hat{\lambda}_{i(s)} }{n_i-1-r}\quad (r=1,...,n_i-2;\ i=1,2).
\label{4.2}
\end{equation}
Note that $\tilde{\lambda}_{i(r)} \ge 0$ w.p.1 for $r=1,...,n_i-2$ and the second term in (\ref{4.2}) is an estimator of 
$\sum_{r=k_i+1}^{p}\lambda_{i(r)}/(n_i-1)\ (=\kappa_i,\ \mbox{say})$.
When applying the NR methodology to the PC direction vector, one obtains 
\begin{equation}
\tilde{\bh}_{i(r)}=\{(n_i-1)\tilde{\lambda}_{i(r)}\}^{-1/2}(\bX_i-\overline{\bX}_i)\hat{\bu}_{i(r)} \ \mbox{ for $r=1,...,n_i-2;\ i=1,2$}.
\label{4.3}
\end{equation}
For $(\hat{\lambda}_{i(r)},\hat{\bh}_{i(r)})$s and $(\tilde{\lambda}_{i(r)},\tilde{\bh}_{i(r)})$s, 
\citet{Aoshima:2016} gave the following results. 
\begin{proposition}[Aoshima and Yata, 2018]
Assume (A-i) and (M-i). 
It holds as $m \to \infty$
\begin{align*}
&\frac{\hat{\lambda}_{i(r)}}{\lambda_{i(r)}}=1+\frac{\kappa_i}{\lambda_{i(r)}}+O_P(n_i^{-1/2}), \ \ (\bh_{i(r)}^T\hat{\bh}_{i(r)})^2=\Big(1+\frac{\kappa_i}{\lambda_{i(r)}}\Big)^{-1}+O_P(n_i^{-1/2}), \\
&\frac{\tilde{\lambda}_{i(r)}}{\lambda_{i(r)}}=1+O_P(n_i^{-1/2}) \ \ {and} \ \ (\bh_{i(r)}^T\tilde{\bh}_{i(r)})^2=1+O_P(n_i^{-1}) 
\end{align*}
for $r=1,...,k_i;\ i=1,2$.
\end{proposition}

If $\kappa_i/\lambda_{i(r)}\to \infty$ as $m \to \infty$, 
$\hat{\lambda}_{i(r)}$ and $\hat{\bh}_{i(r)}$ are strongly inconsistent in the sense that 
$\lambda_{i(r)}/\hat{\lambda}_{i(r)}=o_P(1)$ and $\bh_{i(r)}^T\hat{\bh}_{i(r)}=o_P(1)$. 
For example, in (S-i), $\kappa_i/\lambda_{i(2)}\to \infty$ as $m \to \infty$, 
so that $\bh_{i(2)}^T\hat{\bh}_{i(2)}=o_P(1)$. 
This is the main reason why the classifier by (\ref{3.5}) gave an inadequate performance in Fig. \ref{fig:3}.
On the other hand, $\tilde{\lambda}_{i(r)}$ and $\tilde{\bh}_{i(r)}$ are consistent estimators even when $\kappa_i/\lambda_{i(r)}\to \infty$ as $m \to \infty$. 
We note that $\tr(\bS_i)=\tr(\bSig_i)\{1+o_P(1)\}$ as $m \to \infty$ for $i=1,2$, under (A-i) and (M-i) from the fact that $\Var\{\tr(\bS_i)\}=O\{\tr(\bSig_i^2)/n_i\}=o\{\tr(\bSig_i)^2 \}$ under (A-i) and (M-i). 
Hence, from Proposition 1 we claim that as $m \to \infty$ 
\begin{equation}
\hat{\varepsilon}_{i(r)}=\varepsilon_{i(r)}\{1+o_P(1)\} \ \mbox{ for $r=1,...,k_i;\ i=1,2$, }
\label{nr}
\end{equation}
under (A-i) and (M-i).

Next, we consider an estimation of $x_{0,i(r)}$.
Let
\begin{equation}
\tilde{x}_{0,i(r)}=\bx_{0}^T\tilde{\bh}_{i(r)} \ \mbox{ for all $i,r$.}
\label{4.4}
\end{equation}
Note that Var$({x}_{0,i(r)})=O(\lambda_{i(r)})$ as $p\to \infty$ under (C-vi) when 
$\bx_0 \in \pi_{i'}$ for $r=1,...,k_i;\ i=1,2;\ i'\neq i$. 
Then, we have the following results. 
\begin{proposition}
Assume (A-i), (M-i) and (C-vi). 
Assume also $\limsup_{p\to \infty}$
$[\{\tr(\bSig_{i,A}\bSig_{i'})+\max_{l=1,2}\bmu_{l}^T\bSig_{i,A}\bmu_{l} \}/\lambda_{i(k_i)}^2]<\infty$ 
and  
$ \limsup_{p\to \infty} (\sum_{r=1}^{k_{i}} \{\mu_{i(r)}^2$
$+(\bmu_{i'}^T\bh_{i(r)})^2 \}/\lambda_{i(r)})<\infty$ 
for $i=1,2;\ i'\neq i$. 
Then, 
it holds as $m \to \infty$
\begin{align*}
& \bx_{0}^T\hat{\bh}_{i(r)}=\frac{x_{0,i(r)}}{(1+\kappa_i/\lambda_{i(r)})^{1/2}}+
O_P\{( \lambda_{i(r)}/n_i)^{1/2}\}
 \\
&\mbox{and } \ 
\tilde{x}_{0,i(r)}=x_{0,i(r)}+
O_P\{( \lambda_{i(r)}/n_i)^{1/2}\}
\end{align*}
when $\bx_0 \in \pi_l$ for $r=1,...,k_i;\ i,l=1,2$.
\end{proposition}

Thus one can estimate $x_{0,i(r)}$ by $\tilde{x}_{0,i(r)}$ even when $\kappa_i/\lambda_{i(r)}\to \infty$ as $m \to \infty$.  

Finally, we consider estimating ${x}_{ij(r)}$. 
We note that $\bx_{ij}^T\tilde{\bh}_{i(r)}$ is biased for high-dimensional data. 
This is because $\bx_{ij}^T\tilde{\bh}_{i(r)}$ includes $\|\bx_{ij}-\bmu_i \|^2$ which is very biased for high-dimensional data. 
\CG{Now, we explain the main reason why the inner products involve the large bias terms.  
We note that $\bone_{n_i}^T\hat{\bu}_{i(r)}=0$ and $\bP_{n_i}\hat{\bu}_{i(r)}=\hat{\bu}_{i(r)}$ 
when $\hat{\lambda}_{i(r)}>0$ since $\bone_{n_i}^T\bS_{iD}\bone_{n_i}=0$. 
Also, note that
$$
\{(n_i-1)\tilde{\lambda}_{i(r)}\}^{1/2}\tilde{\bh}_{i(r)}
=\bX_{o,i}\bP_{n_i}\hat{\bu}_{i(r)}=\bX_{o,i}\hat{\bu}_{i(r)} \ \mbox{ when $\hat{\lambda}_{i(r)}>0$,}
$$
where $\bX_{o,i}=\bX_i-\bmu_i\bone_{n_i}^T$. 
Let us write that 
$\hat{\bu}_{i(r)}=(\hat{u}_{i1(r)},...,\hat{u}_{in_i(r)})^T$ for all $i,r$. 
Then, it holds that $\{(n_i-1)\tilde{\lambda}_{i(r)}\}^{1/2}\tilde{\bh}_{i(r)}^T(\bx_{ij}-\bmu_i)=\hat{\bu}_{i(r)}^T\bX_{o,i}^T(\bx_{ij}-\bmu_i)=\hat{u}_{ij(r)}\|\bx_{ij}-\bmu_i \|^2+\sum_{l=1 (\neq j)}^{n_i}\hat{u}_{il(r)}(\bx_{l}-\bmu_i)^T(\bx_{j}-\bmu_i)$, 
so that $\hat{u}_{ij(r)}\|\bx_{ij}-\bmu_i \|^2$ is very biased since $E(\|\bx_{ij}-\bmu_i \|^2)/(n_i-1)\ge \kappa_i$. 
Hence, one should not apply the $\tilde{\bh}_{i(r)}$s (or the $\hat{\bh}_{i(r)}$s) to the estimation of ${x}_{ij(r)}$. 
See Section 5.1 in \citet{Aoshima:2016} for more details.} 
We consider a bias-reduced estimation of ${x}_{ij(r)}$. 
We modify $\hat{\bu}_{i(r)}$ as 
$$
\hat{\bu}_{ij(r)}=(\hat{u}_{i1(r)},...,\hat{u}_{ij-1(r)},-\hat{u}_{ij(r)}/(n_i-1),\hat{u}_{ij+1(r)},...,\hat{u}_{in_i(r)})^T
$$ 
whose $j$-th element is $-\hat{u}_{ij(r)}/(n_i-1)$ for all $i,j,r$. 
Note that 
$\sum_{j=1}^{n_i}\hat{\bu}_{ij(r)}/n_i=\{(n_i-2)/(n_i-1)\}\hat{\bu}_{i(r)}$. 
Let 
\begin{equation}
\tilde{\bh}_{ij(r)}=\frac{(n_i-1)^{1/2}(\bX_i-\overline{\bX}_i)
\hat{\bu}_{ij(r)} }{(n_i-2)\tilde{\lambda}_{i(r)}^{1/2}} \ \mbox{ for all $i,j,r$.}
\notag
\end{equation}
Then, it holds that $\sum_{j=1}^{n_i}\tilde{\bh}_{ij(r)}/n_i=\tilde{\bh}_{i(r)}$ and 
\begin{align*}
&(n_i-2) \{\tilde{\lambda}_{i(r)}/(n_i-1) \}^{1/2}
\tilde{\bh}_{ij(r)}^T
(\bx_{ij}-\bmu_i) \\
&=
(\bx_{ij}-\bmu_i)^T
\bX_{o,i}\bP_{n_i}\hat{\bu}_{ij(r)} 
=\sum_{l=1 (\neq j)}^{n_i} 
\Big(\hat{u}_{il(r)}+\frac{\hat{u}_{ij(r)}}{n_i-1} \Big)(\bx_{ij}-\bmu_i)^T(\bx_{il}-\bmu_i)
\end{align*} 
when $\hat{\lambda}_{i(r)}>0$
from the fact that 
$$
\bP_{n_i}\hat{\bu}_{ij(r)}=
(\hat{u}_{i1(r)},...,\hat{u}_{ij-1(r)},0,\hat{u}_{ij+1(r)},...,\hat{u}_{in_i(r)})^T+(n_i-1)^{-1}\hat{u}_{ij(r)}\bone_{n_i(j)},
$$
where $\bone_{n_i(j)}=(1,...,1,0,1,...,1)^T$ whose $j$-th element is $0$. 
Thus the large biased term, $\|\bx_{ij}-\bmu_i \|^2$, is removed.
Let 
\begin{equation}
\tilde{x}_{ij(r)}=\bx_{ij}^T\tilde{\bh}_{ij(r)} \ \mbox{ for all $i,j,r$.}
\label{4.5}
\end{equation}
See Section 5.1 in \citet{Aoshima:2016} for theoretical comparisons between 
$\bx_{ij}^T\hat{\bh}_{i(r)}$, $\bx_{ij}^T\tilde{\bh}_{i(r)}$ and $\tilde{x}_{ij(r)}$. 
\subsection{Distance-based classifier by the NR methodology}
\label{Sec4.2}
Let $ \overline{\tilde{x}}_{i(r)}=\sum_{j=1}^{n_i}\tilde{x}_{ij(r)}/n_i$ 
for all $i,r$. 
By combining (\ref{4.1}) with (\ref{4.3}), (\ref{4.4}) and (\ref{4.5}), we propose the following classifier: 
\begin{align}
\widetilde{W}_{A}(\bx_0)=&
W(\bx_0)+\sum_{r=1}^{k_1}\tilde{x}_{0,1(r)}\Big\{  \overline{\tilde{x}}_{1(r)}-\frac{1}{2}\tilde{\bh}_{1(r)}^T\Big( \overline{\bx}_{2}- \sum_{s=1}^{k_2}\overline{\tilde{x}}_{2(s)}
\tilde{\bh}_{2(s)}\Big) \Big\}
\notag \\
&-\sum_{r=1}^{k_2} \tilde{x}_{0,2(r)}\Big\{ \overline{\tilde{x}}_{2(r)}
-\frac{1}{2} \tilde{\bh}_{2(r)}^T\Big( \overline{\bx}_{1}- \sum_{s=1}^{k_1} \overline{\tilde{x}}_{1(s)} \tilde{\bh}_{1(s)}\Big)\Big\} \notag \\
&
-\sum_{r=1}^{k_1} \frac{\sum_{j<j'}^{n_1}\tilde{x}_{1j(r)} \tilde{x}_{1j'(r)}}{n_1(n_1-1)}+
\sum_{r=1}^{k_2} \frac{\sum_{j<j'}^{n_2}\tilde{x}_{2j(r)}\tilde{x}_{2j'(r)} }{n_2(n_2-1)}.
 \label{new}
\end{align}
Then, one classifies $\bx_0$ into $\pi_1$ if $\widetilde{W}_{A}(\bx_0)<0$ and into $\pi_2$ otherwise. 
In general, $k_i$s are unknown in $\widetilde{W}_{A}(\bx_0)$.
See Section 4.3 for estimation of $k_i$s. 
\CG{We call the classification rule (\ref{new}) 
the ``transformed distance-based discriminant analysis (T-DBDA)".} 

Now, we give asymptotic properties of T-DBDA. 
We have the following results. 
\begin{theorem}
\label{theorem5}
Assume (A-i) and (M-i). 
Assume also (C-i) to (C-iii) and (C-vi) to (C-viii). 
Then, it holds that as $m\to \infty$
$$
\frac{\widetilde{W}_{A}(\bx_0)}{\Delta_A}=  \frac{(-1)^{i}}{2}+o_P(1) \ \mbox{ when $\bx_0\in\pi_i$ for $i=1,2$.}
$$
For T-DBDA, we have \CG{the classification consistency (\ref{con})} as $m\to \infty$.
\end{theorem}
\begin{theorem}
\label{thm6}
Assume (A-i) and (M-i). 
Assume also (C-iv) to (C-vii) and (C-ix). 
Then, it holds that as $m\to \infty$ 
$$
\frac{\widetilde{W}_A(\bx_0)-(-1)^i\Delta_{A}/2}{\delta_{oi,A} }\Rightarrow N(0,1) \ \mbox{ when $\bx_0\in\pi_i$ for $i=1,2$}
$$
Furthermore, for T-DBDA, (\ref{asy2}) holds as $m\to \infty$. 
\end{theorem}
\begin{remark}
\label{rem5}
From (C-viii) or (C-ix) \CG{T-DBDA} depends on the scale of $\bmu_i$s in the sense that 
$\bmu_{i,A}^T\bSig_{l,A}\bmu_{i,A}$ for $i,l=1,2$. 
Hence, we recommend that one should apply the classifier to a mean-centered data in actual data analyses. 
See Section 5.2 for example.
\end{remark}

\CG{In Fig. \ref{fig:3}, as expected theoretically, 
we observed that $\overline{e}(i)$ for T-DBDA becomes close to that for T-DBDA(b) when $p$ and $n$ are large. 
}
\subsection{Estimation of $k_i$s}
\label{Sec4.3}
In this section, we introduce an estimation of $k_i$ given by \citet{Aoshima:2016}.

Let $n_{i1}=\lceil  n_i/2 \rceil$ and $n_{i2}=n_i-n_{i1}$.
Let $\bX_{i1}=[\bx_{i1},...,\bx_{in_{i1}}]$ and $\bX_{i2}=[\bx_{in_{i1}+1},...,\bx_{in_i}]$ for $i=1,2$. 
We define 
$$
\bS_{iD(1)}=\{(n_{i1}-1)(n_{i2}-1)\}^{-1/2}(\bX_{i1}-\overline{\bX}_{i1})^T(\bX_{i2}-\overline{\bX}_{i2}) \ \ \mbox{for $i=1,2$,}
$$ 
where 
$\overline{\bX}_{il}=[\overline{\bx}_{il},...,\overline{\bx}_{il}]$
with $\overline{\bx}_{i1}=\sum_{j=1}^{n_{i1}}\bx_{ij}/n_{i1}$ 
and $\overline{\bx}_{i2}=\sum_{j=n_{i1}+1}^{n_{i}}\bx_{ij}/n_{i2}$.
Note that rank($\bS_{iD(1)})\le n_{i2}-1$.
By using the cross-data-matrix (CDM) methodology by \citet{Yata:2010}, we estimate 
$\lambda_{i(r)}$ by the $r$-th singular value, $\acute{\lambda}_{i(r)}$, of $\bS_{iD(1)}$, where $\acute{\lambda}_{i(1)}\ge \cdots \ge \acute{\lambda}_{i(n_{i2}-1)}\ge 0$.
\citet{Yata:2010,Yata:2013} showed that $\acute{\lambda}_{i(r)}$ has several consistency properties for high-dimensional non-Gaussian data.
\citet{Aoshima:2011} applied the CDM methodology to obtaining an unbiased estimator of $\tr(\bSig_i^2)$ as $\tr(\bS_{iD(1)}\bS_{iD(1)}^T)$, $i=1,2$. 
Note that $E\{\tr(\bS_{iD(1)}\bS_{iD(1)}^T)\}=\tr(\bSig_i^2)$. 
Also, note that $\acute{\lambda}_{i(r)}^2$ is the $r$-th eigenvalue of $\bS_{iD(1)}\bS_{iD(1)}^T$.
By using the CDM methodology, we consider an estimation of ${\Psi}_{i(r)}$ as $\widehat{\Psi}_{i(1)}=\tr(\bS_{iD(1)}\bS_{iD(1)}^T)$ and 
\begin{equation}
\widehat{\Psi}_{i(r)}=\tr(\bS_{iD(1)}\bS_{iD(1)}^T)-\sum_{s=1}^{r-1}\acute{\lambda}_{i(s)}^2\ \ \mbox{for $r=2,...,n_{i2};\ i=1,2$}.
\label{4.6}
\end{equation}
Note that $\widehat{\Psi}_{i(r)}\ge 0$ w.p.1 for $r=1,...,n_{i2}$, and 
$\hat{\eta}_{i(r)}\in (0,1]$ for $\acute{\lambda}_{i(r)}>0$.
Then, \citet{Aoshima:2016} gave the following result.
\begin{lemma}[Aoshima and Yata, 2018]
\label{lemma1}
Assume (A-i) and (M-i). 
Then, it holds that $\widehat{\Psi}_{i(r)}/{{\Psi}_{i(r)}}=1+o_P(1)$ as $m\to \infty$ for $r=1,...,k_i+1;\ i=1,2$.
\end{lemma}

From (S7.1) in Appendix C of \citet{Aoshima:2016}, it holds that 
$\acute{\lambda}_{i(r)}/\lambda_{i(r)}=1+o_P(1)$ as $m \to \infty$ for $r=1,...,k_i;\ i=1,2$, under (A-i) and (M-i). 
From Lemma \ref{lemma1} we claim under (A-i) and (M-i) that as $m \to \infty$
\begin{equation}
\hat{\eta}_{i(r)}=\eta_{i(r)}\{1+o_P(1)\}
 \ \mbox{ for $r=1,...,k_i;\ i=1,2$.}
\label{cdm}
\end{equation}
Let $\hat{\tau }_{i(r)}=\widehat{\Psi}_{i(r+1)}/\widehat{\Psi}_{i(r)}\ (=1-\acute{\lambda}_{i(r)}^2/\widehat{\Psi}_{i(r)})$ for all $i,r$. 
Note that $1-\hat{\tau }_{i(1)}=\hat{\eta}_{i(1)}$ and $\hat{\tau }_{i(r)}\in [0,1)$ for $\acute{\lambda}_{i(r)}>0$. 
Then, \citet{Aoshima:2016} gave the following result.
\begin{proposition}[Aoshima and Yata, 2018]
\label{pro3}
Assume (A-i) and (M-i). 
It holds for $i=1,2$, that as $m\to \infty$
\begin{align*}
&P(\hat{\tau }_{i(r)}<1-c_r)\to 1 \ \ \mbox{with some fixed constant $c_r\in(0,1)$ for $r=1,...,k_i$};\\
&
\hat{\tau }_{i(k_i+1)}=1+o_P(1).
\end{align*}
\end{proposition}
From Proposition \ref{pro3}, one may choose $k_i$ as the first integer $r$ such that $1-\hat{\tau }_{i(r+1)}$ is sufficiently small. 
In addition, \citet{Aoshima:2016} gave the following result for $\hat{\tau }_{i(k_i+1)}$.
\begin{proposition}[Aoshima and Yata, 2018]
\label{pro4}
Assume (A-i) and (M-i). 
Assume also $\lambda_{i(1)}^2/\Psi_{i(k_i+1)}=o(n_i)$ and 
$\lambda_{i(k_i+1)}^2/\Psi_{i(k_i+1)}=O(n_i^{-c})$ as $m\to \infty$ 
with some fixed constant $c>1/2$ for $i=1,2$. 
It holds for $i=1,2$ that as $m\to \infty$
$$
P\Big(\hat{\tau }_{i(k_i+1)}> \{1+(k_i+1)\gamma(n_i)\}^{-1} \Big) \to 1,
$$
where $\gamma(n_i)$ is a function such that $\gamma(n_i)\to 0$ and $n_i^{1/2}\gamma(n_i)\to \infty$ as $n_i\to \infty$.  
\end{proposition}

From Propositions \ref{pro3} and \ref{pro4}, 
if one can assume the conditions in Proposition \ref{pro4}, 
one may consider $k_i$ as the first integer $r\ (=\hat{k}_{oi},\ \mbox{say})$ such that 
\begin{equation}
\hat{\tau }_{i(r+1)}\{1+(r+1)\gamma(n_i)\}>1 \quad (r \ge 0). \label{4.7}
\end{equation}
Then, it holds that $P(\hat{k}_{oi}=k_i)\to 1$ as $m\to \infty$. 
Note that $\widehat{\Psi}_{i(n_{i2})}=0$ from the fact that rank$(\bS_{iD(1)})\le n_{i2}-1$. 
Thus one may choose $k_i$ as $\hat{k}_{i}=\min\{\hat{k}_{oi},n_{i2}-2\}$ in actual data analyses. 
\citet{Aoshima:2016} recommended to use $\gamma(n_i)=(n_i^{-1} \log{n_i})^{1/2}$. 
Hence, in this paper, we use $\gamma(n_i)=(n_i^{-1} \log{n_i})^{1/2}$ in (\ref{4.7}). 
If $\hat{k}_i=0$ (that is, (\ref{4.7}) holds when $r=0$) for some $i$, 
one may consider the classifier by (\ref{new}) with $\bA_{i}=\bI_p$. 
In addition, if $\hat{k}_i=0$ for $i=1,2$, we recommend to use \CG{DBDA (the classifier by (\ref{1.2}))} because one may assume the NSSE model when $\hat{k}_i=0$ for $i=1,2$. 
\CG{
We summarized $\hat{k}_i$s in Table \ref{tab:1} for the six well-known microarray data sets (D-i) to (D-vi).} 
\section{Performances of the new classifier for the SSE model}
\label{Sec5}
In this section, we discuss the performance of \CG{T-DBDA} 
in numerical simulations and actual data analyses. 
\subsection{Simulation}
\label{Sec5.1}
We compared the performance of \CG{T-DBDA} with other classifiers in complex settings. 
In general, $k_i$s are unknown in (\ref{new}). 
Hence, we estimated $k_i$ by $\hat{k}_{i}$, where $\hat{k}_{i}$ is given in Section 4.3. 
\CG{Hereafter, we describe the classification rule (\ref{new}) with $\hat{k}_{i}$ instead of $k_i$ as ``T-DBDA($*$)".} 
We set $\gamma(n_i)=(n_i^{-1} \log{n_i})^{1/2}$ in (\ref{4.7}).
We set $p=2^{s},\ s=6,...,11$, $\bmu_1=\bze$ and 
$\bmu_{2}=(0,...,0,1,...,1,-1....,-1)^T$ whose last $2 \lceil p^{3/5}/2 \rceil$ elements are not $0$.
The last $\lceil p^{3/5}/2 \rceil$ elements are $-1$ and the previous $\lceil p^{3/5}/2 \rceil$ elements are $1$. 
Note that $\Delta=p^{3/5}\{1+o(1)\}$ as $p\to \infty$. 

First, we considered an intraclass correlation model given by
$$
\bGamma_t=(\bI_t+\bone_{t}\bone_t^T)/2.
$$
Note that $\lambda_{\max}(\bGamma_t)=(t+1)/2$ and the other eigenvalues are $1/2$. 
Let $\bOme_{t}(\rho)=\bB( \rho^{|i-j|^{1/3}})\bB$, where  
$\bB=\mbox{diag}[\{0.5+1/(t+1)\}^{1/2},...,\{0.5+t/(t+1)\}^{1/2}]$. 
Also, note that $[\lambda_{\max}\{\bOme_{t}(\rho)\}]^2$
$/\tr[\{\bOme_{t}(\rho)\}^2]=o(1)$ as $t\to \infty$ for $|\rho|<1$. 
We set $n_1=\lceil p^{1/2} \rceil$, $n_2=2n_1$ and 
\begin{equation}
\bSig_{i}=\left( \begin{array}{ccc}
\bGamma_{p_{i(1)}} &\bO   & \bO \\
\bO & \bGamma_{p_{i(2)}} &\bO  \\
\bO & \bO & c_i\bOme_{p_{i(3)}}(\rho)
\end{array} \right),\  \ i=1,2,
\label{5.1}
\end{equation}
where \CG{$\rho=0.3$}, $p=p_{i(1)}+p_{i(2)}+p_{i(3)}$ and $(c_1,c_2)=(1,1.3)$. 
We considered the following settings:
\begin{description}
\item[(S-ii)] \ 
We generated $\bx_{ij}$, $j=1,2,...\ (i=1,2)$ independently from $N_p(\bmu_i, \bSigma_i)$. 
We set $(p_{1(1)},p_{1(2)})=(\lceil p^{2/3} \rceil,\lceil p^{1/2} \rceil)$ 
and $(p_{1(1)},p_{1(2)})=(2\lceil p^{2/3} \rceil, 2\lceil p^{1/2} \rceil)$;
\item[(S-iii)] \ 
We generated $\bx_{ij}$, $j=1,2,...\ (i=1,2)$ independently from $z_{ij(r)}=(y_{ij(r)}-1)/{2}^{1/2}$  $(r=1,...,p)$ 
in which $y_{ij(r)}$s are i.i.d. as the chi-squared distribution with $1$ degree of freedom. 
We set $(p_{1(1)},p_{1(2)})=(\lceil p/3 \rceil,\lceil p/9 \rceil)$ 
and $(p_{1(1)},p_{1(2)})=(2\lceil p/3 \rceil, 2\lceil p/9 \rceil)$.
\end{description}
For (S-ii) and (S-iii) we note that $\Delta_A=\Delta$ and $\lambda_{i(r)}=(p_{i(r)}+1)/2$, $i,r=1,2,$ for sufficiently large $p$, 
so that (M-i) with $k_1=k_2=2$ is met. 
In particular, the SSSE model (given by (\ref{SSSE})) holds for (S-iii). 
Also, we note that (A-i), (AY-i), (C-i) to (C-iii) and (C-vi) to (C-viii) are met both for (S-ii) and (S-iii), and (AY-ii) is met for (S-ii). 
However, (AY-ii) is not met for (S-iii). 

Next, we considered a Gaussian mixture model whose probability density function is given by
\begin{align}
f_i(\by)=\frac{1}{3} \sum_{l=1}^3 g(\by;\ \bmu_{il(y)},\bSigma_{i(y)}),\ \ i=1,2, \label{5.2}
\end{align}
where $g(\by;\  \bmu_{il(y)},\bSigma_{i(y)})$ is the probability density function of 
$N_p(\bmu_{il(y)},\bSigma_{i(y)})$.
We set $\bSigma_{1(y)}=\bOme_{p}(0.3)$ and $\bSigma_{2(y)}=\bOme_{p}(0.5)$.
Let $q_{1(1)}= \lceil p^{2/3} \rceil$, $q_{2(1)}= 2\lceil p^{2/3} \rceil$, $q_{1(2)}=2\lceil p^{1/2} \rceil$ and 
$q_{2(2)}=\lceil p^{1/2} \rceil$. 
We set 
$\bmu_{i1(y)}=(3^{1/2},...,3^{1/2},0,...,$
$0)^T$ whose first  $q_{i(1)}$ elements are $3^{1/2}$, 
$\bmu_{i2(y)}=(0,...,0,3^{1/2},...,3^{1/2},0,...,0)^T$ whose $(q_{i(1)}+1)$-th to $(q_{i(1)}+q_{i(2)})$-th elements are $3^{1/2}$ and $\bmu_{i3(y)}=\bze$.
We generated $\by_{ij}$, $j=1,2,...\ (i=1,2)$ independently from (\ref{5.2}). 
Note that $E(\by_{ij})=\sum_{l=1}^3\bmu_{il(y)}/3$ for $i=1,2$.
We set $\bx_{ij}=\by_{ij}-\sum_{l=1}^3\bmu_{il(y)}/3+\bmu_i$ for all $i,j$.  
Note that $\bSig_i=\Var(\by_{ij})$ for $i=1,2$, where 
$$
\Var(\by_{ij})=\frac{1}{9}\sum_{l<l'}^3(\bmu_{il(y)}-\bmu_{il'(y)})(\bmu_{il(y)}-\bmu_{il'(y)})^T+\bSigma_{i(y)}.
$$
We note that $\lambda_{i(1)}=(2/3) q_{i(1)}\{1+o(1)\}$ and $\lambda_{i(2)}=(1/2) q_{i(2)}\{1+o(1)\}$
as $p\to \infty$ for $i=1,2$, so that (M-i) with $k_1=k_2=2$ is met. 
See Corollary 2 in \citet{Yata:2015} for the details of the eigenvalues. 
Also, note that $\Delta_A=\Delta$ for sufficiently large $p$ and (A-i) is not met. 
We considered the following settings:
\begin{description}
\item[(S-iv)] \ 
$n_1=\lceil p^{2/5} \rceil$ and $n_2=2n_1$; 
\item[(S-v)\ ] \ 
$n_1=\lceil p^{3/5} \rceil$ and $n_2=2n_1$.
\end{description}
We note that (AY-i), (AY-ii), (C-i) to (C-iii) and (C-vi) to (C-viii) are met both for (S-iv) and (S-v). 

\begin{figure}
\includegraphics[scale=0.55]{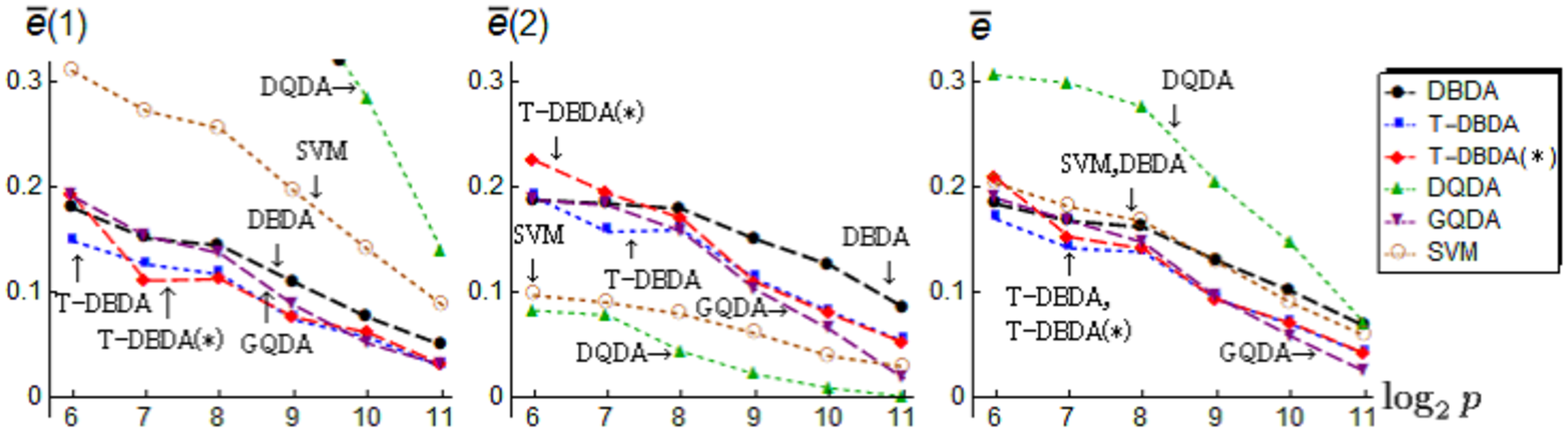}\\
(S-ii): $N_p(\bmu_i, \bSigma_i)$, $(\lambda_{1(1)},\lambda_{1(2)})\approx (p^{2/3}/2,p^{1/2}/2)$ 
and $(\lambda_{2(1)},\lambda_{2(2)})\approx (p^{2/3},p^{1/2})$.
\\[3mm]
\includegraphics[scale=0.55]{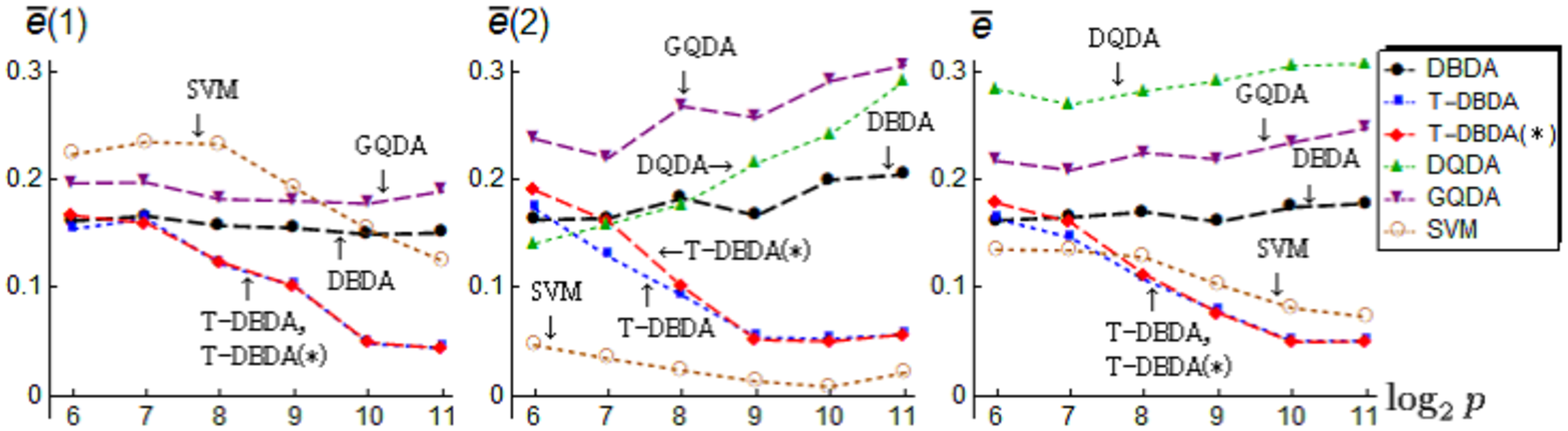}\\
(S-iii): $z_{ij(r)}=(y_{ij(r)}-1)/{2}^{1/2}$  $(r=1,...,p)$ 
in which $y_{ij(r)}$s are i.i.d. as the chi-squared distribution with $1$ degree of freedom, 
$(\lambda_{1(1)},\lambda_{1(2)})\approx (p/6,p/18)$ 
and $(\lambda_{2(1)},\lambda_{2(2)})\approx (p/3,p/9)$.
\\[3mm]
\includegraphics[scale=0.55]{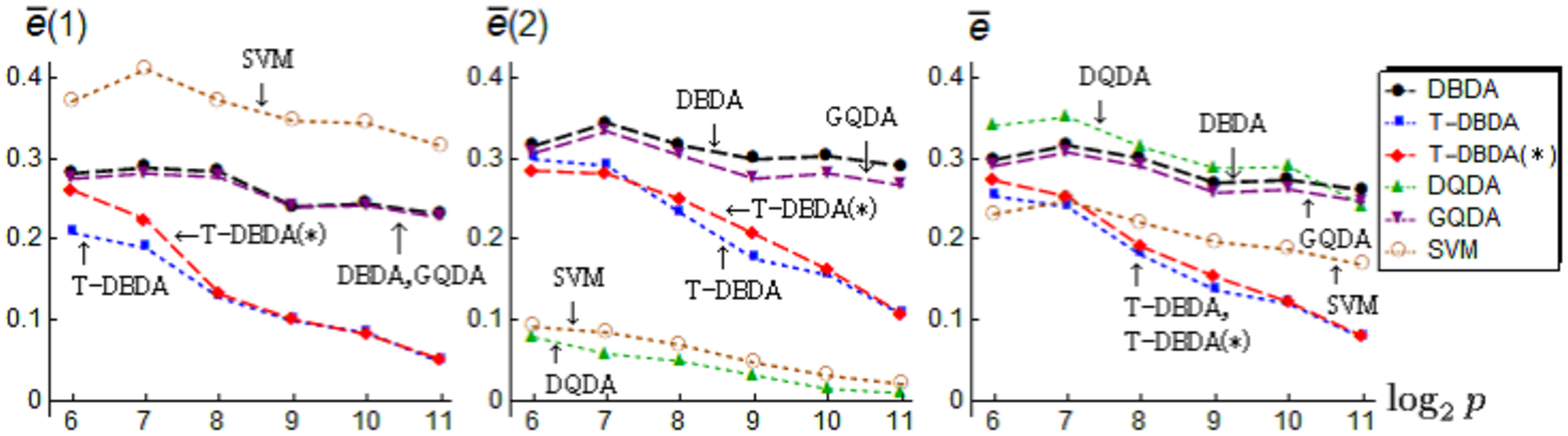}
\\
(S-iv): The mixture model given by (\ref{5.2}) and $(n_1,n_2)=(\lceil p^{2/5} \rceil,2\lceil p^{2/5} \rceil)$.  
\\[3mm]
\includegraphics[scale=0.55]{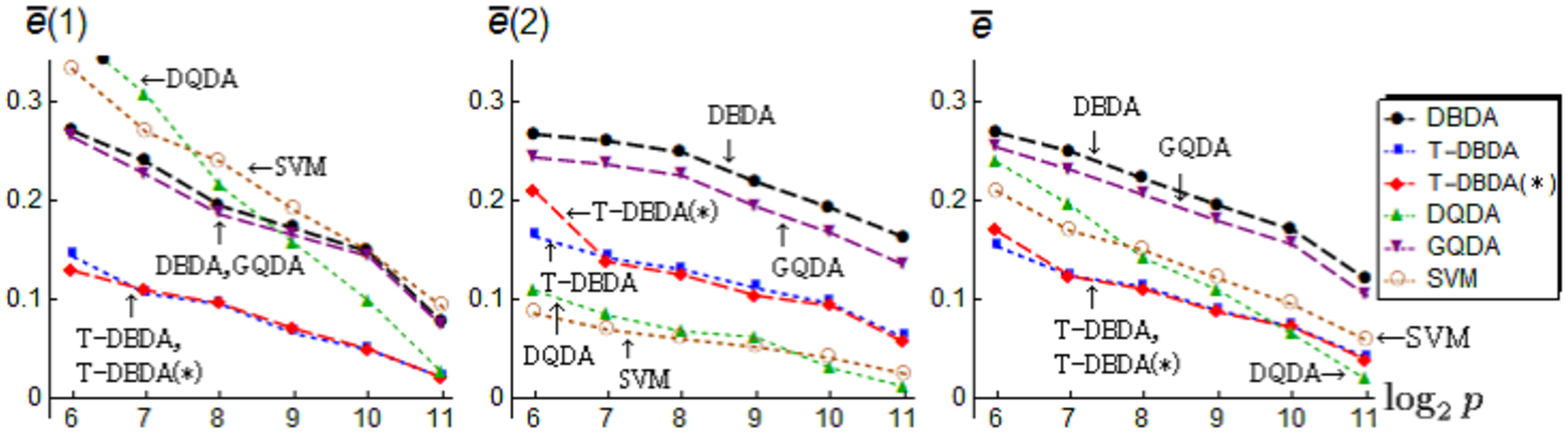}
\\
(S-v): The mixture model given by (\ref{5.2}) and $(n_1,n_2)=(\lceil p^{3/5} \rceil,2\lceil p^{3/5} \rceil)$. 
\caption{
The left panel displays $\overline{e}(1)$, 
the middle panel displays $\overline{e}(2)$ and the right panel displays $\overline{e}$. 
The error rates of the classifiers, DBDA, T-DBDA, T-DBDA($*$), DQDA, GQDA, SVM.
In the left panels, 
$\overline{e}(1)$s for DQDA are not described because the error rates were too high.
}
\label{fig:4} 
\end{figure}

We considered \CG{
DBDA (the classifier (\ref{1.2})), T-DBDA (the classifier (\ref{new})) and T-DBDA($*$) (the classifier (\ref{new}) with $\hat{k}_{i}$ instead of $k_i$). 
We also considered the following three classifiers: 
Diagonal quadratic discriminant analysis (DQDA) given by \citet{Dudoit:2002}, 
Geometrical quadratic discriminant analysis (GQDA) given by \citet{Aoshima:2011,Aoshima:2014}, and Support vector machine (SVM). 
The rule of GQDA is given by (6) in \citet{Aoshima:2014}. 
SVM is the hard-margin linear rule.} 
Similar to Fig. \ref{fig:3}, we calculated the error rates, $\overline{e}(1)$ and $\overline{e}(2)$, by 2000 replications. 
Also, we calculated the average error rate, $\overline{e}=\{\overline{e}(1)+\overline{e}(2)\}/2$. 
Their standard deviations are less than $0.011$. 
In Fig. \ref{fig:4}, we plotted the results for (S-ii) to (S-v). 

We observed that \CG{GQDA gives a better performance compared to DBDA, DQDA and SVM for (S-ii).} 
This is probably because $\tr(\bSig_1)\neq \tr(\bSig_2)$. 
\CG{DQDA performs better compared to DBDA, GQDA and SVM for (S-v). 
}
This is probably because $n_i$s are relatively large and the diagonal elements of the two covariance matrices are not common. 
See Sections 2 to 4 in \citet{Aoshima:2015b} for the details of DQDA and GQDA. 
\CG{For SVM, $\overline{e}(1)$ and $\overline{e}(2)$ were unbalanced. 
The main reason must be due to a bias term in SVM. 
See Section 2 in \citet{Nakayama:2017} for the details.} 
On the other hand, 
DBDA gave a moderate performance for (S-iii). 
This is probably because \CG{DBDA} is quite robust for non-Gaussian HDLSS data. 
See \citet{Aoshima:2014} for the details. 
On the whole, \CG{T-DBDA and T-DBDA($*$) gave adequate performances. 
In particular,  T-DBDA($*$) (or  T-DBDA)} gave a much better performance compared to the other classifiers both for (S-iii), in which (\ref{SSSE}) holds, and (S-iv), in which $n_i$s are relatively small. 
This is probably due to the sufficient conditions of the consistency properties. 
See Section 3.3 for the details. 
The performances of \CG{T-DBDA and T-DBDA($*$)} became quite similar to each other in almost all the cases. 
Hence, we recommend to use ``the classifier (\ref{new}) with $\hat{k}_{i}$ instead of $k_i$" when 
\CG{the SSE condition (\ref{SSE})} or \CG{the SSSE condition (\ref{SSSE})} holds. 
\subsection{Example}
\label{Sec5.2}
In this section, we check the performance of T-DBDA($*$) by using the six well-known microarray data sets in Table \ref{tab:1}. 

First, we used (D-v): myeloma data ($p=12625$). 
We defined $n_1=36$ samples from $\pi_1$ and $n_2=136$ (the first 136) samples from $\pi_2$ as the training data, and the last (the 137-th) sample of $\pi_2$ as the test data. 
We centered each sample by $\bx_{ij}-(\sum_{i'=1}^2\sum_{j'=1}^{n_{i'}}\bx_{i'j'})/(n_1+n_2)$ for all $i,j$, 
and $\bx_0-(\sum_{i'=1}^2\sum_{j'=1}^{n_{i'}}\bx_{i'j'})/(n_1+n_2)$, 
so that $\sum_{i=1}^2\sum_{j=1}^{n_{i}}\bx_{ij}=\bze$. 
We set $\gamma(n_i)=(n_i^{-1} \log{n_i})^{1/2}$ in (\ref{4.7}). 
Let $\tilde{\tau }_{i(r)}=\hat{\tau }_{i(r)}\{1+r\gamma(n_i)\}$ for all $i,r$. 
We calculated that $(\tilde{\tau }_{1(1)},\tilde{\tau }_{1(2)})=(0.943, 1.046)$ 
and $(\tilde{\tau }_{2(1)},\tilde{\tau }_{2(2)},\tilde{\tau }_{2(3)})=(0.878, 0.986, 1.168)$, so that $\hat{k}_{1}=1$ and $\hat{k}_{2}=2$.
Thus, we chose $k_1=1$ and $k_2=2$. 
We calculated that $\widetilde{W}_A(\bx_0)=305.439$, so that we classified $\bx_0$ into $\pi_2$ (the true class). 

Similarly, we checked the accuracy of T-DBDA($*$) by the leave-one-out cross-validation (LOOCV) for (D-i) to (D-vi). 
Also, we checked the accuracy of the classifiers, DBDA, DQDA, GQDA, SVM, by the LOOCV for (D-i) to (D-vi). 
\CG{In addition, we checked the accuracy of the well-known classifiers, 
Diagonal linear discriminant analysis (DLDA) given by \citet{Dudoit:2002} and distance weighted discrimination (DWD) 
given by \cite{Marron:2007}. 
For DWD, we calculated the normal vector by the SOCP solver in \cite{Marron:2007} and set the intercept term as 0 since we used the mean-centered data.}
We summarized misclassification rates, $\overline{e}(1)$, $\overline{e}(2)$ and $\overline{e}=\{\overline{e}(1)+\overline{e}(2)\}/2$, in Table \ref{tab:2}.
\begin{table}
\caption{Error rates of the classifiers by the LOOCV for samples from (D-i) to (D-vi)}
\label{tab:2}       
\begin{tabular}{cccccccc}
\hline
Classifier  & T-DBDA($*$) &  DBDA  &  DLDA & DQDA & GQDA & SVM & DWD   \\
\hline\\[-2mm]
Error rates &  \\[1mm]
\multicolumn{7}{c}{ \ $\pi_1$: 104 samples and $\pi_2$: 113 samples in (D-i)}   \\[1mm]
$\bar{e}_1$  &  0.0    &  0.183  & 0.163  & 0.0  & 0.0 &0.0&0.0   \\
$\bar{e}_2$  &  0.009 &  0.009  & 0.009 & 0.018 &0.044  &0.0&0.009    \\
$\bar{e}$    &  0.004 &  0.096  & 0.086 & 0.009 & 0.022 &0.0&0.004    \\
\hline\\[-2.5mm]
\multicolumn{7}{c}{ \ $\pi_1$: 40 samples and $\pi_2$: 22 samples in (D-ii)}   \\[1mm]
$\bar{e}_1$  &  0.15 &  0.15   & 0.15  & 0.15  & 0.15&0.15&0.15   \\
$\bar{e}_2$  &  0.136 & 0.136  & 0.136 & 0.182 &0.136&0.227&0.091    \\
$\bar{e}$    &  0.143 & 0.143  & 0.143 & 0.166 &0.143&0.189&0.12    \\
\hline\\[-2.5mm]
\multicolumn{7}{c}{\quad \ $\pi_1:$ 111 samples and $\pi_2:$ 57 samples in (D-iii)}   \\[1mm]
$\bar{e}_1$  & 0.198 & 0.243   &0.162 & 0.216 & 0.198 &0.135 &0.243       \\
$\bar{e}_2$  & 0.281   & 0.316   &0.368 & 0.456 & 0.404 &0.439&0.246   \\
$\bar{e}$    & 0.239 & 0.28      &0.265 & 0.336 & 0.301 &0.287&0.244     \\
\hline\\[-2.5mm]
\multicolumn{7}{c}{\quad \ $\pi_1:$ 58 samples and $\pi_2:$ 19 samples in (D-iv)}   \\[1mm]
$\bar{e}_1$  & 0.034 & 0.172   & 0.19 & 0.155 & 0.172&0.017&0.224        \\
$\bar{e}_2$  & 0.0   & 0.158   &0.211 & 0.421 & 0.158&0.0  &0.0  \\
$\bar{e}$    & 0.017 & 0.165   &0.2   & 0.288 & 0.165&0.009&0.112     \\
\hline\\[-2.5mm]
\multicolumn{7}{c}{\quad \ $\pi_1:$ 36 samples and $\pi_2:$ 137 samples in (D-v)}   \\[1mm]
$\bar{e}_1$  & 0.25   & 0.278   &  0.528 &0.639 & 0.278&0.75&0.222  \\
$\bar{e}_2$  & 0.197  & 0.292   & 0.219 & 0.109 & 0.299&0.058&0.365    \\
$\bar{e}$  & 0.224  & 0.285     & 0.373 & 0.374 & 0.289&0.404&0.294      \\
\hline\\[-2.5mm]
\multicolumn{6}{c}{\quad \ $\pi_1:$ 84 samples and $\pi_2:$ 44 samples in (D-vi)}   \\[1mm]
$\bar{e}_1$  & 0.143 & 0.107   & 0.06   &  0.083 &0.143&0.06& 0.107    \\
$\bar{e}_2$  & 0.182 & 0.25    & 0.318  &  0.227 &0.227&0.25 &0.205    \\
$\bar{e}$  & 0.162 & 0.179     & 0.189  &  0.155 &0.185&0.155&0.156  \\
\hline
\end{tabular}
\end{table}

We observed that T-DBDA($*$) gives adequate performances. 
In particular, the new classifier gave a much better performance compared to the other classifiers (except SVM) for (D-iv). 
This is probably because \CG{(D-iv) is close to the SSSE asymptotic domain (\ref{SSSE})}.
See Table \ref{tab:1} or Fig. \ref{fig:1}. 
The other classifiers were probably affected by the strongly spiked eigenvalues directly. 
On the other hand, the new classifier is not directly affected by such eigenvalues. 
See Theorems \ref{thm3} and \ref{theorem5} for the details. 
\CG{This is the reason why the new classifier gave a good performance for (D-iv). 
On the other hand, (D-i) is close to the SSSE asymptotic domain (\ref{SSSE}).
However, the several classifiers gave adequate performances for (D-i). 
This is probably because $n_i$s are relativity large compared to $p$.} 

\section{Proofs}
\label{Sec6}
\subsection{Proof of Theorem 3}
\label{Sec6.1}
We note that for $i,l=1,2;\ i'\neq i$
\begin{equation}
\tr(\bSig_{i,A*}\bSig_{l,A})
=\{\tr(\bSig_{i,A}\bSig_{l,A})+2\tr(\bSig_{i,A}\bSig_{l,A}\bA_{i'} )
+\tr(\bSig_{i}\bA_{i'} \bSig_{l,A}\bA_{i'})\}/4.
\label{A.1}
\end{equation}
From the fact that $\tr(\bSig_{i}\bA_{i'} \bSig_{i,A}\bA_{i'})=\tr(\bSig_{i}^{1/2}\bA_{i'} \bSig_{i,A}\bA_{i'}\bSig_{i}^{1/2})\ge 0\ (i'\neq i)$, 
under (C-ii), 
it holds that $\tr(\bSig_{i,A}^2)/(n_i \Delta_A^2)\to 0$ as $m\to \infty$ for $i=1,2$. 
Thus we claim that 
$
\delta_{oi,A}^2/\Delta_A^2=o(1)
$
for $i=1,2$, under (C-ii).
Note that for $i=1,2$,
\begin{align}
&\bmu_{i}^T\bA_{1,2}\bSig_{l,A}\bA_{1,2}\bmu_i/n_l \le \bmu_{i}^T\bA_{1,2}^2 \bmu_i \lambda_{\max}(\bSig_{l,A})/n_l \notag \\
&\hspace{3.45cm} = (\bmu_{i}^T\bA_{1,2}^2\bmu_i/n_l^{1/2})(\lambda_{l(k_l+1)}/n_l^{1/2}),\ \ l=1,2;\ \mbox{ and} \notag \\
& |\bmu_A^T \bSig_{i',A}\bA_{1,2} \bmu_i| \le \{( \bmu_A^T \bSig_{i',A} \bmu_A)(\bmu_{i}^T\bA_{1,2}\bSig_{i',A}\bA_{1,2}\bmu_i)\}^{1/2},\ \ i'\neq i
 . \label{A.2}
\end{align}
Thus by noting that $\lambda_{l(k_l+1)}=o\{\tr(\bSig_{l,A}^2)^{1/2}\}$ under (M-i) and $\delta_{oi,A}^2/\Delta_A^2=o(1)$ under (C-ii), 
we claim that 
$
\delta_{i,A}^2/\Delta_A^2=o(1)
$
for $i=1,2$, under (M-i), (C-i) to (C-iii). From (\ref{3.2}) and Chebyshev's inequality, we can conclude the results of Theorem 3.  
\subsection{Proof of Corollary 1}
\label{Sec6.2}
By noting that $\tr(\bSig_{i,A*}\bSig_{l,A}) \le \{\tr(\bSig_{i,A}^2)\tr(\bSig_{l,A}^2)\}^{1/2}$ for $i,l=1,2$, when $\bA_1=\bA_2$, the result is obtained straightforwardly from Theorem 3. 
\subsection{Proof of Theorem 4}
\label{Sec6.3}
We first consider the case when $\bx_0\in \pi_1$. 
Let $\omega_{i,A}=\{\tr(\bSig_{i,A*}\bSig_{i,A})/n_i+\tr( \bSig_{i,A*}\bSig_{i',A})/n_{i'}\}^{1/2}$ for $i=1,2;\ i'\neq i$. 
Then, from (\ref{A.1}), under (C-iv), we have that
\begin{equation}
\delta_{o1,A}=\omega_{1,A} \{1+o(1)\} \label{A.3}
\end{equation}
and $\sum_{l=1}^2\tr(\bSig_{l,A}^2)/n_l=O(\delta_{o1,A}^2)$ as $m\to \infty$. 
From (\ref{A.1}), we note that $\lambda_{l(k_l+1)}/n_l^{1/2}=o[\{\tr(\bSig_{l,A}^2)/n_l\}^{1/2}]=o(\delta_{o1,A})$ for $l=1,2$, under (M-i) and (C-iv). 
Thus from (\ref{A.2}) it holds that for $i=1,2$,
\begin{equation}
\delta_{1,A}=\delta_{o1,A}\{1+o(1)\}
\label{A.4}
\end{equation}
under (M-i), (C-iv) and (C-v). 
By combining (\ref{A.3}) and (\ref{A.4}), under (M-i), (C-iv) and (C-v), we have that 
$\delta_{1,A}=\omega_{1,A}\{1+o(1)\}$ and 
\begin{align}
W_{A}(\bx_0)+\frac{\Delta_A}{2}=(\bx_{0}-\bmu_1)^T\bA_*\{(\overline{\bx}_{2,A}-\bmu_{2,A})-(\overline{\bx}_{1,A}-\bmu_{1,A})\}+o_P(\omega_{1,A}).
 \label{A.5}
\end{align}
Let us write that 
\begin{align*}
&v_{j}=-(\bx_0-\bmu_1)^T\bA_*(\bx_{1j,A}-\bmu_{1,A})/(n_1\omega_{1,A}),\quad j=1,...,n_1;\\
&v_{n_1+j}= (\bx_0-\bmu_1)^T\bA_*(\bx_{2j,A}-\bmu_{2,A})/(n_2\omega_{1,A}),\quad j=1,...,n_2.
\end{align*}
Note that $\sum_{j=1}^{n_1+n_2} E(v_{j}^2)=1$ and $\sum_{j=1}^{n_1+n_2}v_{j}=(\bx_{0}-\bmu_1)^T\bA_*\{(\overline{\bx}_{2,A}-\bmu_{2,A})-(\overline{\bx}_{1,A}-\bmu_{1,A})\}/\omega_{1,A}$.
Then, it holds that $E(v_{j}|v_{j-1},...,v_{1})=0$ for $j=2,...,n_1+n_2$.
We consider applying the martingale central limit theorem given by \citet{McLeish:1974}.
In a way similar to the equations (23) and (24) in \citet{Aoshima:2014}, we can evaluate that under (A-i)
\begin{align}
& (n_{l_j} \omega_{1})^4E(v_{j}^4)=O[\tr(\bSig_{1,A*} \bSig_{l_j,A})^2+\tr\{(\bSig_{1,A*} \bSig_{l_j,A})^2 \}] \ \ \mbox{and}
\label{A.6} \\
&(n_{l_j}n_{l_{j'}})^2 \omega_{1}^4 E(v_{j}^2v_{j'}^2) \notag\\
&=\tr(\bSig_{1,A*} \bSig_{l_j,A})\tr(\bSig_{1,A*} \bSig_{l_{j'},A})
+O\{\tr(\bSig_{1,A*} \bSig_{l_j,A}\bSig_{1,A*} \bSig_{l_{j'},A} )\} \notag \\
&\hspace{0.4cm} +O[\{\tr(\bSig_{1,A*} \bSig_{l_j,A}\bSig_{1,A*} \bSig_{l_j,A})
\tr(\bSig_{1,A*} \bSig_{l_{j'},A}\bSig_{1,A*} \bSig_{l_{j'},A})\}^{1/2}]\label{A.7} 
\end{align}
for $j\neq j'$, 
where $l_j=1$ for $j \in [1,...,n_1]$ and $l_j=2$ for $j\in [n_1+1,...,n_1+n_2]$.
For any $\tau>0$ we note that 
$\sum_{j=1}^{n_1+n_2} E \{  v_{j}^2 I ( v_{j}^2 \ge \tau  ) \}\le \sum_{j=1}^{n_1+n_2}E(v_{j}^4)/\tau$ 
from Chebyshev's inequality and Schwarz's inequality, 
where $I(\cdot)$ is the indicator function. 
Also, note that $\tr\{(\bSig_{1,A*} \bSig_{l,A})^2 \}\le \tr(\bSig_{1,A*} \bSig_{l,A})^2$ for $l=1,2$. 
Then, from (\ref{A.6}), under (A-i), it holds that for Lindeberg's condition 
$$
\sum_{j=1}^{n_1+n_2} E \{  v_{j}^2 I ( 
 v_{j}^2 \ge \tau  ) \}=O\Big[\frac{\tr(\bSig_{1,A*} \bSig_{1,A})^2/n_1^3 +\tr(\bSig_{1,A*} \bSig_{2,A})^2/n_2^3}{
 \omega_{1,A}^4 } \Big]=o(1)
$$
for any $\tau>0$. 
Note that for $l,l'=1,2$,
\begin{align*}
\tr(\bSig_{1,A*} \bSig_{l,A}\bSig_{1,A*} \bSig_{l',A})
&=\tr\{(\bSigma_{1,A*}^{1/2} \bSig_{l,A}\bSigma_{1,A*}^{1/2} )(\bSig_{1,A*}^{1/2} \bSigma_{l',A} \bSig_{1,A*}^{1/2})\} \\
&\le \lambda_{\max}(\bSig_{1,A*}^{1/2}\bSigma_{l,A} \bSig_{1,A*}^{1/2})\tr(\bSig_{1,A*}^{1/2} \bSigma_{l',A} \bSig_{1,A*}^{1/2}) 
\\
&=o\{\tr(\bSig_{1,A*} \bSig_{l,A})\tr(\bSig_{1,A*} \bSig_{l',A})\}
\end{align*}
under (C-v), so that 
$
(n_{l_j}n_{l_{j'}})^2 \omega_{1}^4 E(v_{j}^2v_{j'}^2)=
\tr(\bSig_{1,A*} \bSig_{l_j,A})\tr(\bSig_{1,A*} \bSig_{l_{j'},A})\{1+o(1)\}
$
for $j\neq j'$. 
Hence, 
by using Chebyshev's inequality, from (\ref{A.6}) and (\ref{A.7}), under (A-i) and (C-v), it holds that for any $\tau>0$ 
$$
P\Big(\Big|\sum_{j=1}^{n_1+n_2} v_{j}^2-1\Big|\ge \tau \Big)\le \frac{E[\sum_{j,j'=1}^{n_1+n_2} 
\{v_{j}^2-E(v_{j}^2)\}\{v_{j'}^2-E(v_{j'}^2)\}] }{\tau^2}= o(1),
$$
so that $ \sum_{j=1}^{n_1+n_2} v_{j}^2=1+o_P(1)$. 
Hence, by using the martingale central limit theorem, 
we obtain that $\sum_{j=1}^{n_1+n_2}v_j \Rightarrow N(0,1)$ under (A-i) and (C-v). 
Thus from (\ref{A.5}) we conclude the result when $\bx_0\in \pi_1$. 
When $\bx_0\in \pi_2$, we can conclude the result similarly.
The proof is completed. 
\subsection{Proof of Corollary 2}
\label{Sec6.4}
When $\bA_1=\bA_2$,
we note that $\lambda_{\max}(\bSigma_{i,A*}^{1/2}  \bSig_{l,A} \bSigma_{i,A*}^{1/2})\le \lambda_{i(k_i+1)}
\lambda_{l(k_l+1)}$ 
and $\tr(\bSigma_{i,A*}\bSigma_{l,A})=\tr(\bSigma_{i,A}\bSigma_{l,A})$ for $i,l=1,2$. 
On the other hand, when $\bA_1=\bA_2$, it holds that 
$
\bmu_A^T\bSigma_{i',A}\bmu_A/(n_{i'} \delta_{oi,A}^2)=o(1) 
$
as $m\to \infty$ for $i=1,2;\ i'\neq i$, 
under $\bmu_A^T \bSigma_{i',A} \bmu_A/(\delta_{oi',A}^2)=o(1)$ as $m\to \infty$ 
and $\tr(\bSigma_{1,A}^2)/\tr(\bSigma_{2,A}^2)\in(0,\infty)$ as $p\to \infty$.
Hence, from Theorem 4 we can conclude the results.
\subsection{Proof of Proposition 2}
\label{Sec6.5}
We assume (A-i) and (M-i).
Let $\bu_{i(r)}=(z_{i1(r)},...,z_{in_i(r)})/(n_i-1)^{1/2}$
and $\dot{\bu}_{i(r)}={\|\bu_{i(r)}\|}^{-1}{{\bu}_{i(r)}}$ for all $i,j$.
Then, from (S6.1) to (S6.3) and (S6.5) in Appendix B of \citet{Aoshima:2016}, we can claim that as $m\to \infty$ for $i=1,2$,
\begin{align}
&\tilde{\lambda}_{i(r)}/\lambda_{i(r)}=||\bu_{i(r)}||^2+
O_P(n_i^{-1})=1+O_P(n_i^{-1/2})\notag \\
&\mbox{and} \ \ \hat{\bu}_{i(r)}^T\dot{\bu}_{i(r)}=1+O_P(n_i^{-1})
\ \  \mbox{for $r=1,...,k_i$;}
 \label{A.11} \\
&\hat{\bu}_{i(s)}^T{\bu}_{i(r)}= O_P(n_i^{-1/2}\lambda_{i(s)}/\lambda_{i(r)})\notag \\
&\mbox{and} \ \ \hat{\bu}_{i(r)}^T{\bu}_{i(s)}= O_P(n_i^{-1/2})
\ \ \mbox{for $r<s \le k_i$}.\label{A.12}
\end{align}
From (\ref{A.11}) there exists a unit random vector $\bzeta_{i(r)}$ such that $\dot{\bu}_{i(r)}^T\bzeta_{i(r)}=0$ and 
\begin{align}
\hat{\bu}_{i(r)}=\{1+O_P(n_i^{-1})\}\dot{\bu}_{i(r)}+\bzeta_{i(r)}\times O_P(n_i^{-1/2})  \label{A.13}
\end{align}
for $r=1,...,k_i;\ i=1,2$. 
We note that $\bone_{n}^T\hat{\bu}_{i(r)}=0$ and $\bP_{n_i}\hat{\bu}_{i(r)}=\hat{\bu}_{i(r)}$ 
when $\hat{\lambda}_{i(r)}>0$ since $\bone_{n_i}^T\bS_{iD}\bone_{n_i}=0$. 
Also, when $\hat{\lambda}_{i(r)}>0$, note that
\begin{align*}
\tilde{\bh}_{i(r)}
=\frac{(\bX_i-\bmu_i\bone_{n_i}^T)\bP_{n_i}\hat{\bu}_{i(r)}}{\{(n_i-1)\tilde{\lambda}_{i(r)}\}^{1/2}} 
=\frac{\sum_{s=1}^p\lambda_{i(s)}^{1/2}\bh_{i(s)}\bu_{i(s)}^T  \hat{\bu}_{i(r)}}{\tilde{\lambda}_{i(r)}^{1/2}},
\end{align*}
so that $\bx_0^T\tilde{\bh}_{i(r)}=\sum_{s=1}^p\lambda_{i(s)}^{1/2}x_{0,i(s)}\bu_{i(s)}^T\hat{\bu}_{i(r)}/\tilde{\lambda}_{i(r)}^{1/2}$. 
Here, we claim that when $\bx_0\in \pi_l,\ l=1,2$,
\begin{align*}
&E\Big\{\Big(\frac{\sum_{s=k_i+1}^p  \lambda_{i(s)}^{1/2}x_{0,i(s)} \bu_{i(s)}^T{\bu}_{i(r)}}{\lambda_{i(r)}^{1/2}}\Big)^2\Big\}=O\Big\{\frac{\tr( \bSig_{l} \bSig_{i,A})+\bmu_{l}^T\bSig_{i,A}\bmu_{l}  }{n_i\lambda_{i(r)}} \Big\};\\
&E\Big\{\Big\|\frac{\sum_{s=k_i+1}^p  \lambda_{i(s)}^{1/2}x_{0,i(s)} \bu_{i(s)}}{\lambda_{i(r)}^{1/2}}\Big\|^2\Big\}=O\Big\{\frac{\tr( \bSig_{l} \bSig_{i,A})+\bmu_{l}^T\bSig_{i,A}\bmu_{l}  }{\lambda_{i(r)}} \Big\}
\end{align*}
for $r=1,...,k_i;\ i=1,2$. 
Then, from (\ref{A.11}) and (\ref{A.13}), it holds that when 
$\bx_0\in \pi_l,\ l=1,2$,
\begin{equation}
\frac{\sum_{s=k_i+1}^p  \lambda_{i(s)}^{1/2}x_{0,i(s)} \bu_{i(s)}^T\hat{\bu}_{i(r)}}{\tilde{\lambda}_{i(r)}^{1/2}}=O_P\Big\{\Big( \frac{\tr( \bSig_{l} \bSig_{i,A})+\bmu_{l}^T\bSig_{i,A}\bmu_{l}  }{n_i\lambda_{i(r)}}\Big)^{1/2} \Big\}
\label{A.14}
\end{equation}
for $r=1,...,k_i;\ i=1,2$, from the fact that 
$ \sum_{s=k_i+1}^p  \lambda_{i(s)}^{1/2} x_{0,i(s)} \bu_{i(s)}^T \bzeta_{i(r)}$
$/\lambda_{i(r)}^{1/2}\le 
\|\lambda_{i(r)}^{-1/2}$
$ \sum_{s=k_i+1}^p  \lambda_{i(s)}^{1/2}x_{0,i(s)} \bu_{i(s)} \|\cdot \|\bzeta_{i(r)} \|$ and Markov's inequality. 
Note that $E(x_{0,i(s)}^2)=\bh_{i(s)}^T(\bSig_l+\bmu_l\bmu_l^T)\bh_{i(s)}$ 
when $\bx_0\in \pi_l\ (l=1,2)$ for all $i,s$, so that $x_{0,i(s)}=O_P[\{ \bh_{i(s)}^T(\bSig_l+\bmu_l\bmu_l^T)\bh_{i(s)}\}^{1/2}]$.
Then, from (\ref{A.11}) and (\ref{A.12}), 
we have that when $\bx_0\in \pi_l,\ l=1,2$,
\begin{align}
&\frac{\sum_{s=1}^{k_i}  \lambda_{i(s)}^{1/2}x_{0,i(s)} \bu_{i(s)}^T\hat{\bu}_{i(r)}}{\tilde{\lambda}_{i(r)}^{1/2}}
\notag \\
&=x_{0,i(r)}+
O_P\Big\{ \Big(\sum_{s=1}^{k_i}\frac{\lambda_{i(s)} \bh_{i(s)}^T(\bSig_l+\bmu_l\bmu_l^T)\bh_{i(s)}
  }{n_i\max\{\lambda_{i(s)}^2/\lambda_{i(r)},\lambda_{i(r)} \}}\Big)^{1/2} \Big\}
\label{A.15}
\end{align}
for $r=1,...,k_i;\ i=1,2$. 
By combining (\ref{A.14}) and (\ref{A.15}), we can conclude the second result of Proposition 2. 
For the first result, 
from Proposition 1 and the second result, it concludes the result.
\subsection{Proofs of Theorems 5 and 6}
\label{Sec6.6}
Assume (A-i) and (M-i). 
We first consider the proof of Theorem 5. 
Let $\psi_{i(r)}=\tr(\bSig_{i}^2)/(n_i^2\lambda_{i(r)})+\bmu_i^T\bSig_i\bmu_i/(n_i\lambda_{i(r)})$ 
for $r=1,...,k_i;\ i=1,2$. 
Then, from Lemma B.1 and (S6.27) in Appendix B of \citet{Aoshima:2016}, 
we claim that as $m\to \infty$
\begin{equation}
\overline{\tilde{x}}_{i(r)}=\bar{x}_{i(r)}+O_P(\psi_{i(r)}^{1/2}) \ \ 
\mbox{and} \ \ 
\bar{x}_{i(r)}=\mu_{i(r)}+O_P\{(\lambda_{i(r)}/n_i)^{1/2}\} 
\label{A.16}
\end{equation}
for $r=1,...,k_i;\ i=1,2$.
Note that under (C-vii) 
\begin{align}
\psi_{i(r)}=O\Big(\frac{\lambda_{i(1)}^2+n_{i}\bmu_{i,A}^T\bSig_{i,A}\bmu_{i,A} }{n_{i}^2\lambda_{i(r)}}\Big)
\ \ \mbox{for $r=1,...,k_i;\ i=1,2$.}
\label{A.17}
\end{align}
Note that $\tr(\bSig_{i,A}\bSig_{i'})=\tr(\bSig_{1,A}\bSig_{2,A})+O(\lambda_{i(k_i)}\lambda_{i'(1)})=O(\lambda_{i(k_i)}\lambda_{i'(1)})$ 
and
$\bmu_{i'}^T\bSig_{i,A}\bmu_{i'}=O(\bmu_{i',A}^T\bSig_{i,A}\bmu_{i',A}+\sum_{s=1}^{k_{i'}} \lambda_{i(k_i)} \mu_{i'(s)}^2)$
for $i=1,2;\ i'\neq i$ from the facts that $\tr(\bSig_{1,A}\bSig_{2,A})\le \{\tr(\bSig_{1,A}^2)\tr(\bSig_{2,A}^2) \}^{1/2}=O(\lambda_{1(k_1)}\lambda_{2(k_2)})$ and $\bmu_{i',A}^T\bSig_{i,A}\bh_{i'(s)}\mu_{i'(s)}=O(\bmu_{i',A}^T\bSig_{i,A}\bmu_{i',A}+\lambda_{i(k_i)} \mu_{i'(s)}^2)$ for $s=1,...,k_{i'}$. 
From (\ref{A.14}) and (\ref{A.15})
we have that when $\bx_{0}\in \pi_l,\ l=1,2$,
\begin{align}
&\tilde{x}_{0,i(r)}={x}_{0,i(r)}+
O_P\Big\{\Big(\frac{\lambda_{i(r)}^2+\bmu_{l,A}^T\bSig_{i,A}\bmu_{l,A} }{n_i\lambda_{i(r)}}\Big)^{1/2}\Big\}+O_P\{(\lambda_{l(1)}/n_i)^{1/2}\} \notag \\
\mbox{and} \ \ & {x}_{0,i(r)}=O_P(\lambda_{i(r)}^{1/2}) 
\ \ \mbox{for $r=1,...,k_i;\ i=1,2$} \label{A.18}
\end{align}
under (C-vi) and (C-vii). 
Then, from (\ref{A.16}) to (\ref{A.18}), under (C-vi) to (C-viii), 
we have that when $\bx_{0}\in \pi_l,\ l=1,2$,
\begin{align}
\tilde{x}_{0,i(r)}\overline{\tilde{x}}_{i(r)}-
{x}_{0,i(r)}\bar{x}_{i(r)}&=
(\tilde{x}_{0,i(r)}-{x}_{0,i(r)})\overline{\tilde{x}}_{i(r)}
+{x}_{0,i(r)}(\overline{\tilde{x}}_{i(r)}-\bar{x}_{i(r)}) \notag \\
&=o_P(\Delta_A) \ \ \mbox{for $r=1,...,k_i;\ i=1,2$}.
\label{A.19}
\end{align}
On the other hand, from (S6.29) in Appendix B of \citet{Aoshima:2016} 
we claim that for $r=1,...,k_1$ and $s=1,...,k_2$ 
\begin{align}
&\tilde{\bh}_{1(r)}^T
\tilde{\bh}_{2(s)}=
{\bh}_{1(r)}^T{\bh}_{2(s)}+O_P(n_{\min}^{-1/2}),\ \ 
\tilde{\bh}_{1(r)}^T
(\tilde{\bh}_{2(s)}-{\bh}_{2(s)})=O_P(n_2^{-1/2}), \notag \\
&\tilde{\bh}_{2(s)}^T
(\tilde{\bh}_{1(r)}-{\bh}_{1(r)})=O_P(n_1^{-1/2}) \notag \\
&\mbox{and} \ \ 
(\tilde{\bh}_{1(r)}-{\bh}_{1(r)})^T
(\tilde{\bh}_{2(s)}-{\bh}_{2(s)})=O_P\{(n_1n_2)^{-1/2}\}. \label{A.20}
\end{align}
Note that $ \bar{x}_{i(r)}\bh_{i(r)}-\overline{\tilde{x}}_{i(r)}\tilde{\bh}_{i(r)}
=\bar{x}_{i(r)}(\bh_{i(r)}-\tilde{\bh}_{i(r)})-(\overline{\tilde{x}}_{i(r)}-\bar{x}_{i(r)})\tilde{\bh}_{i(r)}$ for all $i,r$. 
Then, 
from (\ref{A.16}) and (\ref{A.20}), we have that for $r=1,...,k_i;\ i=1,2;\ i'\neq i$,
\begin{align}
& \tilde{\bh}_{i(r)}^T \sum_{s=1}^{k_{i'}}( \bar{x}_{i'(s)}\bh_{i'(s)}-\overline{\tilde{x}}_{i'(s)}\tilde{\bh}_{i'(s)}) \notag \\
&=O_P\Big(\sum_{s=1}^{k_{i'}}( \psi_{i'(s)}^{1/2}({\bh}_{i(r)}^T{\bh}_{i'(s)}+n_{\min}^{-1/2})+ \lambda_{i'(s)}^{1/2}/n_{i'}+\mu_{i'(s)}/n_{i'}^{1/2})\Big). 
\label{A.21}
\end{align}
Similar to the proof of Proposition 2 and (\ref{A.18}), under (C-vi) and (C-vii), 
we can claim that for $r=1,...,k_i;\ i=1,2;\ i'\neq i$,
\begin{align}
\tilde{\bh}_{i(r)}^T\overline{\bx}_{i',A}=&
{\bh}_{i(r)}^T\overline{\bx}_{i',A}+
O_P\Big\{\Big( \frac{\lambda_{i(r)}^2/n_{\min}+\bmu_{i',A}^T\bSig_{i,A}\bmu_{i',A}}{n_i\lambda_{i(r)}}\Big)^{1/2} \Big\}
\notag \\
&+O_p[\{\lambda_{i'(1)}/(n_1n_2)\}^{1/2}].
\label{A.22}
\end{align}
Note that $ \sum_{s=1}^{k_{i'}} ({\bh}_{i(r)}^T{\bh}_{i'(s)})^2/\lambda_{i'(s)}=O(1/\lambda_{i(r)})$ 
under (C-vi) for $r=1,...,k_i$; $i=1,2;\ i'\neq i$.
From (\ref{A.17}), (\ref{A.21}) and (\ref{A.22}) we have that for $r=1,...,k_i;\ i=1,2;\ i'\neq i$,
\begin{align}
&\tilde{\bh}_{i(r)}^T
\Big(\overline{\bx}_{i'}-\sum_{s=1}^{k_{i'}}\overline{\tilde{x}}_{i'(s)}\tilde{\bh}_{i'(s)}\Big)
-{\bh}_{i(r)}^T
\Big(\overline{\bx}_{i'}-\sum_{s=1}^{k_{i'}}\bar{x}_{i'(s)}{\bh}_{i'(s)}\Big) \notag \\
&=O_P\Big\{\Big( \frac{\bmu_{i',A}^T\bSig_{i,A}\bmu_{i',A}}{n_i\lambda_{i(r)}}+
\frac{\bmu_{i',A}^T\bSig_{i',A}\bmu_{i',A}}{ \min\{\lambda_{i(r)},n_{\min}\lambda_{i'(k_{i'})} \} n_{i'}}  \notag \\
&\hspace{1.5cm} +
 \frac{\lambda_{i(1)}+\lambda_{i'(1)}}{n_{\min}^2}+\frac{\lambda_{i'(1)}^2}{n_{\min}^2\lambda_{i(r)} }\Big)^{1/2}\Big\}
\label{A.23}
\end{align}
under (C-vi) and (C-vii).
Note that ${\bh}_{i(r)}^T (\overline{\bx}_{i'}-\sum_{s=1}^{k_{i'}}\bar{x}_{i'(s)}{\bh}_{i'(s)})=O_P\{(\lambda_{i(r)}$
$/n_{i'})^{1/2}\}$ under (C-vi) and (C-vii) for $r=1,...,k_i;\ i=1,2;\ i'\neq i$. 
Then, similar to (\ref{A.19}), from (\ref{A.18}) and (\ref{A.23}), 
we have that 
\begin{align}
&\tilde{x}_{0,i(r)}
\tilde{\bh}_{i(r)}^T
\Big(\overline{\bx}_{i'}-\sum_{s=1}^{k_{i'}}\overline{\tilde{x}}_{i'(s)}\tilde{\bh}_{i'(s)}\Big)
-{x}_{0,i(r)}{\bh}_{i(r)}^T
\Big(\overline{\bx}_{i'}-\sum_{s=1}^{k_{i'}}\bar{x}_{i'(s)}{\bh}_{i'(s)}\Big) \notag \\
&=o_P(\Delta_A) \ \ \mbox{for $r=1,...,k_i;\ i=1,2;\ i'\neq i$}
\label{A.24}
\end{align}
under (C-vi) to (C-viii).
Also, from (S6.28) in Appendix B of \citet{Aoshima:2016}, we claim that for $r=1,...,k_i;\ i=1,2$,
\begin{align}
&\sum_{j<j'}^{n_i} \frac{ \tilde{x}_{ij(r)} \tilde{x}_{ij'(r)}-{x}_{ij(r)} {x}_{ij'(r)}}{n_i(n_i-1)}
=O_P\Big\{\psi_{i(r)}^{1/2}(\psi_{i(r)}^{1/2}+\lambda_{i(r)}^{1/2}/n_i^{1/2}+\mu_{i(r)})\Big\}.
\notag
\end{align}
Note that under (C-vii) and (C-viii)
\begin{align}
&\sum_{r=1}^{k_i}\psi_{i(r)}^{1/2}(\psi_{i(r)}^{1/2}+\lambda_{i(r)}^{1/2}/n_i^{1/2}+\mu_{i(r)}) \notag \\
&=O\Big(
\frac{\lambda_{i(1)}\lambda_{i(k_i)}+\bmu_{i,A}^T\bSig_{i,A}\bmu_{i,A} }{n_i\lambda_{i(k_i)}} 
+
\frac{(\lambda_{i(1)}^2+n_i\bmu_{i,A}^T\bSig_{i,A}\bmu_{i,A})^{1/2} }{n_i^{3/2}} \Big)=o_P(\Delta_A)
\label{A.25}
\end{align}
for $i=1,2$. By combining (\ref{A.19}), (\ref{A.24}) and (\ref{A.25}), 
it holds that 
$\widetilde{W}_{A}(\bx_0)={W}_{A}(\bx_0)+o_P(\Delta_A)$  
when $\bx_0 \in \pi_i,\ i=1,2$ under (C-vi) to (C-viii).
It concludes the results of Theorem 5. 

Similar to the proof of Theorem 5, it holds that 
$\widetilde{W}_{A}(\bx_0)={W}_{A}(\bx_0)+o_P(\delta_{o\min,A})$ 
when $\bx_0 \in \pi_i,\ i=1,2$ under (C-vi), (C-vii) and (C-ix).
It concludes the results of Theorem 6.

\vskip 14pt
\noindent {\large\bf Acknowledgements}

Research of the first author was partially supported by Grants-in-Aid for Scientific Research (A) and 
Challenging Exploratory Research, Japan Society for the Promotion of Science (JSPS), under Contract Numbers 15H01678 and 26540010. 
Research of the second author was partially supported by Grant-in-Aid for Young Scientists (B), JSPS, under Contract Number 26800078.
\par



\begin{thebibliography}{99}
%




\bibitem[Ahn and Marron(2010)]{Ahn:2010}
Ahn, J., Marron, J.S. (2010).
The maximal data piling direction for discrimination. 
{\em Biometrika, 97}, 254--259.


\bibitem[Alon et al.(1999)]{Alon:1999}
Alon, U., Barkai, N., Notterman, D.A., Gish, K., Ybarra, S., Mack, D., Levine, A.J. (1999).  
Broad patterns of gene expression revealed by clustering analysis of tumor and normal colon tissues probed by oligonucleotide arrays.
{\em Proceedings of the National Academy of Sciences of the United States of America, 96}, 6745--6750.


\bibitem[Aoshima and Yata(2011)]{Aoshima:2011}
Aoshima, M., Yata, K. (2011).
Two-stage procedures for high-dimensional data. 
{\em Sequential Analysis (Editor's special invited paper), 30}, 356--399. 

\bibitem[Aoshima and Yata(2014)]{Aoshima:2014}
Aoshima, M., Yata, K. (2014).
A distance-based, misclassification rate adjusted classifier for multiclass, high-dimensional data.
{\em Annals of the Institute of Statistical Mathematics, 66}, 983--1010. 

\bibitem[Aoshima and Yata(2015a)]{Aoshima:2015a}
Aoshima, M., Yata, K. (2015a).
Geometric classifier for multiclass, high-dimensional data. 
{\em Sequential Analysis, 34}, 279--294. 

\bibitem[Aoshima and Yata(2015b)]{Aoshima:2015b}
Aoshima, M., Yata, K. (2015b).
High-dimensional quadratic classifiers in non-sparse settings. 
{\em arXiv preprint}, arXiv:1503.04549.

\bibitem[Aoshima and Yata(2018)]{Aoshima:2016}
Aoshima, M., Yata, K. (2018).
Two-sample tests for high-dimension, strongly spiked eigenvalue models. 
{\em Statistica Sinica}, to appear (arXiv:1602.02491). 

\bibitem[Bai and Saranadasa(1996)]{Bai:1996}
Bai, Z., Saranadasa, H. (1996).
Effect of high dimension: By an example of a two sample problem. 
{\em Statistica Sinica, 6}, 311--329.


\bibitem[Bickel and Levina(2004)]{Bickel:2004}
Bickel, P.J., Levina, E. (2004).
Some theory for {F}isher's linear discriminant function, ``naive Bayes", and some alternatives when there are many more variables than observations. 
{\em Bernoulli, 10}, 989--1010. 

\bibitem[Cai and Liu(2011)]{Cai:2011}
Cai, T.T., Liu, W. (2011).
A direct estimation approach to sparse linear discriminant analysis. 
{\em Journal of the American Statistical Association, 106}, 1566--1577. 


\bibitem[Chan and Hall(2009)]{Chan:2009}
Chan, Y.-B., Hall, P. (2009).
Scale adjustments for classifiers in high-dimensional, low sample size settings.
{\em Biometrika, 96}, 469--478. 

\bibitem[Chen and Qin(2010)]{Chen:2010}
Chen, S.X., Qin, Y.-L. (2010). 
A two-sample test for high-dimensional data with applications to gene-set testing. 
{\em The Annals of Statistics, 38}, 808--835. 

\bibitem[Christensen et al.(2009)]{Christensen:2009}
Christensen, B.C., Houseman, E.A., Marsit, C.J., Zheng, S., Wrensch, M.R., Wiemels, J.L., Nelson, H.H. et al. (2009). 
Aging and environmental exposures alter tissue-specific DNA methylation dependent upon CpG island context. 
{\em PLoS Genetics, 5}, e1000602. 





\bibitem[Dudoit et al.(2002)]{Dudoit:2002}
Dudoit, S., Fridlyand, J., Speed, T.P. (2002).
Comparison of discrimination methods for the classification of tumors using gene expression data.
{\em Journal of the American Statistical Association, 97}, 77--87. 



\bibitem[Fan and Fan(2008)]{Fan:2008}
Fan, J., Fan, Y. (2008).
High-dimensional classification using features annealed independence rules. 
{\em The Annals of Statistics, 36}, 2605--2637.

\bibitem[Glaab et al.(2012)]{Glaab:2012}
Glaab, E., Bacardit, J., Garibaldi, J.M., Krasnogor, N. (2012). 
Using rule-based machine learning for candidate disease gene prioritization and sample
classification of cancer gene expression data. 
{\em PLoS ONE, 7}, e39932. 


\bibitem[Gravier et al.(2010)]{Gravier:2010}
Gravier, E., Pierron, G., Vincent-Salomon, A., Gruel, N., Raynal, V., Savignoni, A., De Rycke, Y. et al. (2010). 
A prognostic DNA signature for T1T2 node-negative breast cancer patients.
{\em Genes, Chromosomes and Cancer, 49}, 1125--1134.




\bibitem[Hall et al.(2005)]{HMN:2005}
Hall, P., Marron, J.S., Neeman, A. (2005).
Geometric representation of high dimension, low sample size data.
{\em Journal of the Royal Statistical Society, Series B, 67}, 427--444.

\bibitem[Hall et al.(2008)]{Hall:2008}
Hall, P., Pittelkow, Y., Ghosh, M. (2008).
Theoretical measures of relative performance of classifiers for high dimensional data with small sample sizes. 
{\em Journal of the Royal Statistical Society, Series B, 70}, 159--173.

\bibitem[Jeffery et al.(2006)]{Jeffery:2006}
Jeffery, I.B., Higgins, D.G., Culhane, A.C. (2006). 
Comparison and evaluation of methods for generating differentially expressed gene lists from microarray data. 
{\em BMC Bioinformatics 7}, 359.



\bibitem[Li and Shao(2015)]{Li:2015}
Li, Q., Shao, J. (2015). 
Sparse quadratic discriminant analysis for high dimensional data.
{\em Statistica Sinica, 25}, 457--473. 


\bibitem[Marron et al.(2007)]{Marron:2007}
Marron, J.S., Todd, M.J., Ahn, J. (2007).
Distance-weighted discrimination. 
{\em Journal of the American Statistical Association, 102}, 1267--1271. 

\bibitem[McLeish(1974)]{McLeish:1974}
McLeish, D.L. (1974). 
Dependent central limit theorems and invariance principles. 
{\em The Annals of Probability, 2}, 620--628. 


\bibitem[Naderi et al.(2007)]{Naderi:2007}
Naderi, A., Teschendorff, A.E., Barbosa-Morais, N.L., Pinder, S.E., Green, A.R., Powe, D.G., 
Robertson, J.F. et al. (2007). 
A gene-expression signature to predict survival in breast cancer across independent data sets. 
{\em Oncogene, 26}, 1507--1516. 

\bibitem[Nakayama et al.(2017)]{Nakayama:2017}
Nakayama, Y., Yata, K., Aoshima, M. (2017). 
Support vector machine and its bias correction in high-dimension, low-sample-size settings. 
{\em Journal of Statistical Planning and Inference, 191}, 88--100.


\bibitem[Ramey(2016)]{Ramey:2016}
Ramey J.A. (2016). 
Datamicroarray: collection of data sets for classification. 
\url{https://github.com/ramhiser/datamicroarray}. 


\bibitem[Shao et al.(2011)]{Shao:2011}
Shao, J., Wang, Y., Deng, X., Wang, S. (2011). 
Sparse linear discriminant analysis by thresholding for high dimensional data.
{\em The Annals of Statistics, 39}, 1241--1265. 


\bibitem[Shipp et al.(2002)]{Shipp:2002}
Shipp, M.A., Ross, K.N., Tamayo, P., Weng, A.P., Kutok, J.L., Aguiar R.C., Gaasenbeek, M. et al. (2002). 
Diffuse large B-cell lymphoma outcome prediction by gene-expression profiling and supervised machine learning.
{\em Nature Medicine 8}, 68--74. 




\bibitem[Tian et al.(2003)]{Tian:2003}
Tian, E., Zhan, F., Walker, R., Rasmussen, E., Ma, Y., Barlogie, B., Shaughnessy, J.D. Jr. (2003). 
The role of the Wnt-signaling antagonist DKK1 in the development of osteolytic lesions in multiple myeloma. 
{\em The New England Journal of Medicine, 349}, 2483--2494. 


\bibitem[Watanabe et al.(2015)]{Watanabe:2015}
Watanabe, H., Hyodo, M., Seo, T., Pavlenko, T. (2015). 
Asymptotic properties of the misclassification rates for Euclidean Distance Discriminant rule in high-dimensional data. 
{\em  Journal of Multivariate Analysis, 140}, 234--244.

\bibitem[Yata and Aoshima(2010)]{Yata:2010}
Yata, K., Aoshima, M. (2010). 
Effective PCA for high-dimension, low-sample-size data with singular value decomposition of cross data matrix.
{\em  Journal of Multivariate Analysis, 101}, 2060--2077.

\bibitem[Yata and Aoshima(2012)]{Yata:2012}
Yata, K., Aoshima, M. (2012). 
Effective PCA for high-dimension, low-sample-size data with noise reduction via geometric representations. 
{\em  Journal of Multivariate Analysis, 105}, 193--215.

\bibitem[Yata and Aoshima(2013)]{Yata:2013}
Yata, K., Aoshima, M. (2013). 
PCA consistency for the power spiked model in high-dimensional settings.
{\em  Journal of Multivariate Analysis, 122}, 334--354.

\bibitem[Yata and Aoshima(2015)]{Yata:2015}
Yata, K., Aoshima, M. (2015). 
Principal component analysis based clustering for high-dimension, low-sample-size data. 
{\em arXiv preprint}, arXiv:1503.04525.






\end{thebibliography}
\end{document}